\documentclass[onecolumn, draftclsnofoot]{IEEEtran}
\usepackage{ amssymb  }
\usepackage{ mathdots }
\usepackage{ amsmath  }
\usepackage{  color   }
\usepackage{ graphicx }
\usepackage{ amsfonts }
\usepackage{ amssymb  }
\usepackage{ epstopdf }
\usepackage{  upgreek }
\usepackage{    bm    }
\usepackage{ graphicx }
\usepackage{ subfigure}
\usepackage{ mathtools}
\usepackage{ algorithm}
\usepackage{algorithmic}
\usepackage{   cite   }
\usepackage{booktabs}
\usepackage{multirow}
\usepackage{url}

\DeclareMathOperator*{\argmin}{arg\,min}
\DeclareMathOperator*{\argmax}{arg\,max}
\DeclareMathOperator{\Tr}{Tr}

\newcommand{\norm}[1]{\left\lVert#1\right\rVert}

\begin{document}
\IEEEoverridecommandlockouts
\title{ Wireless Traffic Prediction with Scalable Gaussian Process: Framework, Algorithms, and Verification }

\author{\IEEEauthorblockN{Yue Xu$^{\dag*}$, Feng Yin$^{*}$, Wenjun Xu$^{\dag}$,  Jiaru Lin$^{\dag}$, Shuguang Cui$^{\ddagger*}$\bigskip}
	
\IEEEauthorblockA{
	$^{\dag}$Key Lab of Universal Wireless Communications, Ministry of Education \\
	Beijing University of Posts and Telecommunications\\
	$^{*}$SRIBD and The Chinese University of Hong Kong, Shenzhen\\
	$^{\ddagger}$Department of Electrical and Computer Engineering\\
	University of California, Davis\\}
}

\maketitle
\begin{abstract}
	The cloud radio access network (C-RAN) is a promising paradigm to meet the stringent requirements of the fifth generation (5G) wireless systems. Meanwhile, wireless traffic prediction is a key enabler for C-RANs to improve both the spectrum efficiency and energy efficiency through load-aware network managements.
	This paper proposes a scalable Gaussian process (GP) framework as a promising solution to achieve large-scale wireless traffic prediction in a cost-efficient manner. 
	Our contribution is three-fold. 
	First, to the best of our knowledge, this paper is the first to empower GP regression with the alternating direction method of multipliers (ADMM) for parallel hyper-parameter optimization in the training phase, where such a scalable training framework well balances the local estimation in baseband units (BBUs) and information consensus among BBUs in a principled way for large-scale executions. 
	Second, in the prediction phase, we fuse local predictions obtained from the BBUs via a cross-validation based optimal strategy, which demonstrates itself to be reliable and robust for general regression tasks. Moreover, such a cross-validation based optimal fusion strategy is built upon a well acknowledged probabilistic model to retain the valuable closed-form GP inference properties. 
	Third, we propose a C-RAN based scalable wireless prediction architecture, where the prediction accuracy and the time consumption can be balanced by tuning the number of the BBUs according to the real-time system demands.
	Experimental results show that our proposed scalable GP model can outperform the state-of-the-art approaches considerably, in terms of wireless traffic prediction performance.
\end{abstract}


\section{Introduction}
\label{sec:Introduction}
The fifth generation (5G) system is expected to provide approximately 1000 times higher wireless capacity and reduce up to 90 percent of energy consumption compared with the current 4G system~\cite{7064897}. One promising solution to reach such ambitious goals is the adoption of cloud radio access networks (C-RANs)~\cite{C-RAN}, which have attracted intense research interests from both academia and industry in recent years~\cite{7444125}. 
A C-RAN is composed of two parts: the distributed remote radio heads (RRHs) with basic radio functionalities to provide coverage over a large area, and the centralized baseband units (BBUs) pool with parallel BBUs to support joint processing and cooperative network management. 
The BBUs can perform dynamic resource allocation in accordance with real-time network demands based on the virtualized resources in cloud computing. One major feature for the C-RANs to enable high energy-efficient services is the fast adaptability to non-uniform traffic variations~\cite{7444125, C-RAN, 8114545, 7064897}, e.g., the tidal effects. 
Consequently, wireless traffic prediction techniques stand out as the key enabler to realize such load-aware management and proactive control in C-RANs, e.g., the load-aware RRH on/off operation~\cite{8114545}.
However, the adoption of wireless traffic prediction techniques in C-RANs must satisfy the requirements on prediction accuracy, cost-efficiency, implementability, and scalability for large-scale executions.

\subsection{Related Works}
In the literature, many statistical time series models and analysis methods have been proposed for wireless traffic prediction.
For instance, the linear autoregressive integrated moving average (ARIMA) model has been used to model the short-term correlation in network traffic~\cite{10.1007}. However, wireless network traffic often shows a long-term correlation due to mobile user behaviors~\cite{4024193}. 
As an extension, Shu \textit{et al.}~\cite{shu2003wireless} adopted the seasonal ARIMA (SARIMA) models to improve the ARIMA-based models on long-term traffic correlation modeling. 
In~\cite{wang2015approach}, Wang \textit{et al.} proposed the sinusoid superposition model to describe both the short-term and long-term traffic patterns based on a liner combination of different frequency components, which are extracted by performing the fast Fourier transform (FFT) on traffic curves. 

However, wireless network traffic is becoming more complex with increasing non-linear patterns, which are difficult to be captured via linear models. Therefore, machine learning based models could be adopted to improve the prediction accuracy, e.g., those based on deep neural networks. For instance, Nie \textit{et al.}~\cite{7925498} exploited the deep belief network (DBN) and Gaussian model for both the low-pass and high-pass components of the wireless traffic respectively. Qiu \textit{et al.}~\cite{8264694} exploited the recurrent neural network (RNN) to predict wireless traffic based on certain spatial-temporal information. However, (deep) neural networks are also well-known for the difficulty of training, and their learned features packed in a black-box are usually hard to interpretate.
Recently, the Gaussian process (GP) model, which is a class of Bayesian nonparametric machine learning models, has achieved outstanding performance in various fields~\cite{7870565, 7467578}. 
Comparing with other machine learning methods, e.g., the neural networks, the GP model does not involve any black-box operations. Instead, the GP encodes domain/expert knowledge into the kernel function and optimizes the hyper-parameters explicitly based on the Bayes theorems to generate certain explainable results. Therefore, it has a great potential in improving the interpretability and prediction accuracy. Moreover, GP could provide the posterior variance of the predictions; in other words, GP not only predicts the future traffic, but also provides a measure of uncertainty over the predicted results, which is of vital importance for robust network managements, e.g., the network routing based on traffic uncertainty~\cite{6736759}, and the resource reservation for cell on/off operations~\cite{8304392}. In addition, GP is also promising for modeling and analyzing spatial-temporal data~\cite{Senanayake}.

Nevertheless, the bottleneck of standard GP model lies in the high computational complexity, which creates difficulties for large-scale executions in the C-RANs. 
This motivates the researchers to seek for low-complex GP methods that are capable of achieving similar prediction performance as the standard ones but with much lower computational complexity.
Among others, the following two methods are promising for large-scale applications. The first is the sparse GP model with the idea to approximate the distribution of the full dataset based on its subset~\cite{CR05}. However, the selection of such an optimal subset is difficult and time-consuming. The other alternative is the distributed GP model, which splits the heavy training burden to multiple parallel machines and then optimally combine local inferences to generate an improved global prediction. 
Distributed GP models comprise two operation phases in sequence, i.e., the joint hyper-parameter learning in the training phase and the optimal fusion of local inferences in the prediction phase. The existing distributed GP models all come with certain limits in either of the two phases or both.
For example, the Bayesian committee machine (BCM) proposed by Tresp~\cite{Tresp00} approximates the likelihood function over the full dataset with a product of likelihood functions over its subsets.
However, during the BCM's training phase, each local GP model optimizes its own hyper-parameters independently without any interactions, which forbids information sharing from promoting joint improvements. 
The robust BCM (rBCM) proposed by Deisenroth~\cite{deisenroth2015distributed} is based on a product-of-experts (PoE) framework to address the shortcomings of BCM. 
However, in the training phase, rBCM assumes that all local GP models are trained jointly with complete information sharing, which incurs extensive communication overheads in large-scale executions.
Besides, in the prediction phase, rBCM uses heuristic weights to combine the local inferences, which is straightforward but not robust. 

Lastly but importantly, most of the existing literatures on wireless traffic prediction only focus on the algorithm design without considering their adaptation to practical wireless architectures. 
For example, the BBUs are aggregated in a few big rooms and can be selectively turned on or off according to the actual network demands to curtail the operation expenditures, e.g., reducing the power consumption of air conditioning. 
In addition, how to exploit the parallel BBUs for joint processing to improve the prediction speed thus reducing the operation time
and how to accommodate the utilization of BBUs and the scalability of prediction models still remain to be explored.

\subsection{Contributions}
In this paper, we propose a scalable GP-based wireless prediction framework based on a C-RAN enabled architecture. The proposed framework leverages the parallel BBUs in C-RANs to perform predictions with computational complexity depending on the number of active BBUs. Therefore, the prediction accuracy and the time consumption can be well-balanced by activating or deactivating the BBUs, which constitutes a cost-efficient solution for large-scale executions. 
The main contributions of this work are summarized as follows.
\begin{itemize}
	\item To the best of our knowledge, this work is the first to apply GP for wireless traffic prediction with a tailored kernel function that can capture both the periodic trend and the dynamic deviations observed in real 4G data. The obtained prediction accuracy reaches up to 97\%, much higher than the existing methods. Besides, compared with the existing works that are typically based on 2G or 3G wireless traffic data, the proposed models are more promising to be applied in 5G wireless networks. 
	\item This work proposes a C-RAN based wireless prediction architecture to depict a feasible realization with C-RAN infrastructures, where the RRHs collect and deliver the local traffic data to the BBU pool, while the BBU pool performs on-demand wireless traffic predictions by readily changing the number of active BBUs. Such a C-RAN based architecture is promising for joint parallel processing and large-scale cooperative optimization to realize intelligent load-aware network management.
	\item This paper proposes a scalable GP framework based on the distributed GP with significant innovations in both the training phase and the prediction phase. Specifically, 
	1) \textit{for the GP training phase:} this work is the first to propose a scalable training framework based on the alternating direction method of multipliers (ADMM) algorithm. The training framework provides an elegant and principled route for performing hyper-parameter learning and information exchanging among parallel computing units, i.e., BBUs, and moreover achieves excellent tradeoff between the communication overhead and training performance. For each BBU, the computational complexity of training phase can be reduced from $ \mathcal{O}(N^3) $ of the standard GP to $ \mathcal{O}(\frac{N^3}{K^3}) $ of our proposed scalable GP, where $ N $ is the number of training points and $ K $ is the number of parallel BBUs.
	2) \textit{For the GP prediction phase:} this work is the first to fuse the prediction results from local GP models elegantly via optimizing the fusion weights based on a desired performance metric with validation points. We prove that when the validation set only contains a single data point, the weight optimization problem can be cast into a convex problem with a global optimal solution. When the validation set contains more than one data points, the weight optimization problem can be solved via mirror descent efficiently with convergence guarantees. Moreover, we propose a simplified weighting strategy based on the soft-max function for high-speed executions particularly for real-time applications.
	\item In order to further decrease the computational complexity of each BBU, we propose the structured GP model for kernel matrices with a Toeplitz structure. This structure arises when dealing with a dataset with regularly-spaced input grid, concretely, in our case, the wireless traffic recorded with regular time intervals. The structured GP model can further reduce the training complexity from $ \mathcal{O}(\frac{N^3}{K^3}) $ to $ \mathcal{O}(\frac{N^2}{K^2}) $ without sacrificing any prediction accuracy.
\end{itemize}

The remainder of this paper is organized as follows. 
Section~\ref{sec:system model} presents the system model including the C-RAN enabled architecture and the standard GP regression model.
Section~\ref{sec:scalable gp} presents the scalable GP framework, where the ADMM-empowered scalable training framework and cross-validation based scalable fusion framework are established respectively.
Section~\ref{sec:results} demonstrates the experimental results. 
Finally, Section~\ref{sec:conclusion} concludes this paper. 
\section{System Model}
\label{sec:system model}
\subsection{C-RAN based Wireless Traffic Prediction Architecture}
\label{sec:architecture}
\begin{figure}
	\centering
	\includegraphics[trim=10 10 10 10, clip, width=8.5cm]{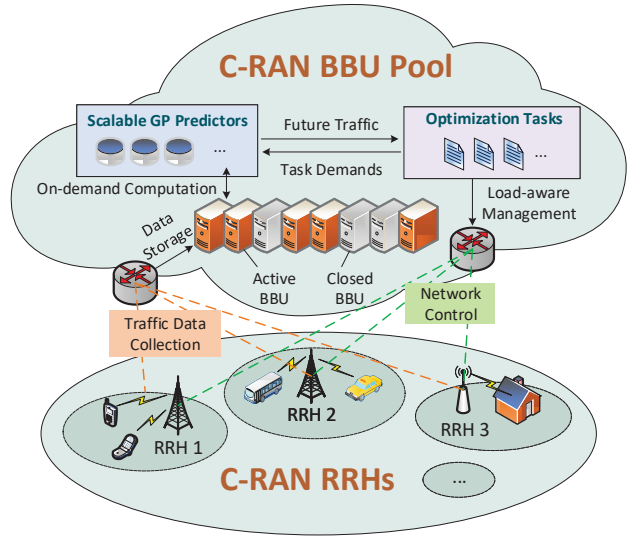}
	\caption{A general architecture for wireless traffic prediction based on C-RANs }
	\label{fig:architechture}
\end{figure}

The C-RAN combines powerful cloud computing and flexible virtualization techniques based on a centralized architecture~\cite{C-RAN, 7444125}, which makes it a decent paradigm to support large-scale wireless predictions with the following advantages.
First, the C-RAN can support on-demand computing resource allocation for cost-efficient service commitments, such as changing the number of active BBUs according to real-time computing demands~\cite{7444125}.
Second, the centralized and virtualized hardware platform can better support internal information sharing, which lays the foundation for joint parallel processing with reduced communication overheads and rapid data assignments.
Third, the centralized C-RAN architecture can also better support cooperative managements~\cite{7875131}, such that the wireless traffic prediction result can directly guide the load-aware network management to improve both the spectrum efficiency and energy efficiency of wireless networks, which contributes a candidate solution to realize adaptive resource allocation and proactive network control for future intelligent management.

Motivated by the above benefits, we propose a C-RAN based wireless traffic prediction architecture as shown in Fig.~\ref{fig:architechture}.
The proposed architecture inherits the two-layer structure, i.e., the RRHs deployed at remote sites, and the BBUs clustered as the BBU pool centrally. The RRHs are equipped with basic radio functions to monitor the local traffic data at different areas, and deliver them to the BBU pool via the common public radio interface (CPRI) protocol~\cite{7444125}, where the BBU pool performs traffic predictions for all RRHs accordingly based on the available BBUs.
To support large-scale and real-time executions, we propose a scalable wireless traffic prediction framework that can well-balance the prediction accuracy and time consumption by readily changing the number of running BBUs. 
Specifically, each traffic prediction model is performed on an individual BBU and trained based on its assigned subset split from the full training dataset. Consequently, higher prediction accuracy can be achieved when using less BBUs but with larger subsets, for that more information can be preserved from dataset division. The extreme case is only using one BBU to learn from the full dataset directly. On the contrary, less time consumption can be achieved by continuously increasing the number of running BBUs with smaller subsets, which however will bring certain prediction performance loss.
In this way, the prediction tasks triggered by different RRHs can be easily matched with appropriate computing resources for the best service commitment. In other words, the C-RAN can selectively activate or deactivate the BBUs according to the actual network demands in terms of, e.g., acceptable accuracy levels, delay requirements, task priorities, system burdens, etc. Therefore, our proposed architecture can dynamically and adaptively balance the energy consumption and prediction performance to provide a cost-effective solution. 

The proposed traffic prediction architecture could greatly assist in traffic-aware managements and adaptive network control to accomplish better network scheduling and resource configuration in future C-RANs.
For example, the proposed architecture could well-guide the RRH on/off management to largely improve the energy efficiency, e.g., the traffic-aware RRH on/off operations in the 5G C-RAN~\cite{7445140}, the load-aware RRH on/off operations in the green heterogeneous C-RAN~\cite{8114545}. On the other hand, the GP-based traffic prediction model has already been applied to guide the base station on/off operations in wireless networks~\cite{8304392}.
Other usage cases for the wireless traffic prediction model include the traffic-aware resources management in the 5G core network~\cite{8553653}, the 5G traffic management under the enhanced mobile broadband (eMBB), massive machine type communications (mMTC), and ultra-reliable low-latency communication (URLLC) scenarios~\cite{8553655}, etc.
Moreover, the proposed C-RAN-based prediction architecture is also able to flexibly control the time consumption over online traffic predictions by readily changing the number of BBUs, which can therefore better support delay-sensitive services and improve the user experience on latency. Meanwhile, since the GP prediction phase requires considerably less time consumption than the GP training phase (which will be discussed in the following sections), the C-RAN can also choose to train the GP model offline, and use the well-trained GP model to perform online predictions in accomplishing ultra-fast traffic prediction responses.\footnote{However, it is suggested to re-train the GP model periodically with the most recent traffic samples in order to retain the prediction accuracy.}
Additionally, it is noteworthy to point out that the proposed scalable GP framework is indeed a general-purpose regression framework, which is likely to find wide applications on other regression tasks by changing the kernel function and the learning context, e.g., building a GP-based fingerprint map with the SE kernel and RSS measurements as the training data~\cite{7870565}.

\subsection{General Patterns of Wireless Traffic Data}
\label{sec:patterns}
We first analyze the general patterns in wireless traffic data as the basis to craft a decent prediction model.
The dataset used in this paper contains hourly recorded downlink physical resource block (PRB) usage\footnote{The PRB usage can reflect wireless traffic flow changes and is therefore chosen to be the prediction target.} histories of about 3000 4G base stations.
\begin{figure}
	\centering
	\includegraphics[trim=20 20 30 20, clip, width=7cm]{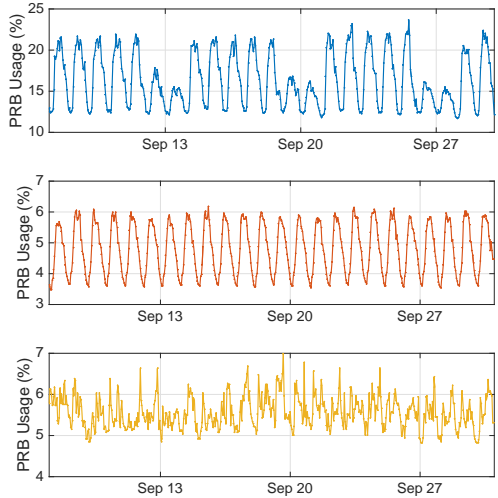}
	\caption{The PRB usage curves of three base stations.}
	\label{fig:typical_traffic_week}
\end{figure}
The three panels of Fig. \ref{fig:typical_traffic_week} show the PRB usage curves of three different base stations. The curve in the first panel represents one base station in an office area, where the traffic trend shows a strong weekly periodic pattern in accordance with weekdays and weekends. The curve in the second panel represents one base station in a residential area, which shows a strong daily pattern with higher demands in the daytime and lower demands in the nighttime. The curve in the third panel represents one base station in a rural area, where the weekly and daily patterns are not obvious. Apart from the periodic patterns, all three curves show irregular dynamic deviations.
To summarize, the 4G wireless traffic in our dataset shows three general patterns: 1) \textit{weekly periodic pattern}: the variations in accordance with weekdays and weekends; 2) \textit{daily periodic pattern}: the variations in accordance with weekdays and weekends; 3) \textit{dynamic deviations}: the variations on top of the above periodic trends.

Generally, the periodic traffic variations are mainly caused by the aggregated user behaviors incorporating the 24-hour periodic nature of human mobility and the periodical tendency to visit the same locations~\cite{5935313}.  
Similar periodicities have already been observed in the 3G traffic~\cite{5935313} and the simulated dense C-RAN system~\cite{7445140}.
Hence, it is highly likely that the 5G traffic will also exhibit similar periodic patterns as long as the nature of human mobility still remains.
Actually, the 4G traffic already exhibits larger dynamic deviations than the 3G traffic due to the diversification of services and the evolution of network architectures, which implies further exaggerated dynamic deviations in the future 5G traffic due to the ongoing network developments.
Therefore, in this paper, we aim to propose a general prediction model that has the flexibility to fit both the periodic patterns and the dynamic deviations with different magnitudes, thereby being widely applicable to general traffic datasets collected from the real 4G networks or the future 5G networks. 

\subsection{GP-based Wireless Traffic Prediction Model}
\subsubsection{GP-based Regression Model}
A Gaussian process is an important class of the kernel-based machine learning methods. It is a collection of random variables, any finite number of which have Gaussian distributions~\cite{RW06}. In this paper, we focus on the real-valued Gaussian processes that are completely specified by a mean function and a kernel function.
Specifically, for the wireless traffic of each RRH in the C-RAN, we consider the following regression model:
\begin{equation}
y = f(\boldsymbol{x}) + e, 
\end{equation}
where $y \in \mathbb{R}^{1}$ is a continuously valued scalar output; $e$ is the independent noise, which is assumed to be Gaussian distributed with zero mean and variance $\sigma_e^2$; $f(\boldsymbol{x})$ is the regression function, which is described with a GP model as
\begin{equation}
f(\boldsymbol{x}) \sim \mathcal{GP}(m(\boldsymbol{x}), k(\boldsymbol{x}, \boldsymbol{x}'; \boldsymbol{\theta}_{h})),
\end{equation} 
where $m(\boldsymbol{x}) $ is the mean function, often assumed to be zero in practice, especially when no prior knowledge is available, and $k(\boldsymbol{x}, \boldsymbol{x}'; \boldsymbol{\theta}_{h})$ is the kernel function determined by the kernel hyper-parameters $\boldsymbol{\theta}_{h}$. Hence, the hyper-parameters to be learned\footnote{Although $\sigma_e^2$ can be estimated jointly with the kernel hyper-parameters $ \boldsymbol{\theta}_{h} $, in this paper we assume it is estimated independently by some other estimation processes, e.g., the robust smoothing method~\cite{Garcia10}.}  for wireless traffic prediction can be denoted as $\boldsymbol{\theta} \triangleq [\boldsymbol{\theta}_{h}^{T}, \sigma_e^2]^T$.


Generally, the prediction task can be summarized as follows. Given a training dataset $\mathcal{D} \triangleq \{\boldsymbol{X}, \boldsymbol{y} \}$, where $\boldsymbol{y} = [y_1, y_2, ..., y_n]^T$ is the training outputs and $\boldsymbol{X}=[\boldsymbol{x}_1, \boldsymbol{x}_2,...,\boldsymbol{x}_n]$ is the training inputs, the aim is to predict the output $\boldsymbol{y}_{*} =  [y_{*,1}, y_{*,2},...,y_{*,n_{*}}]^T$ given the test inputs $\boldsymbol{X}_{*} = [\boldsymbol{x}_{*,1}, \boldsymbol{x}_{*,2},...,\boldsymbol{x}_{*,n_{*}}]$ based on the posterior distribution $p(\boldsymbol{y}_{*} \vert \mathcal{D}, \boldsymbol{X}_{*}; \boldsymbol{\theta})$. 
According to the definition of Gaussian process given beforehand, the joint prior distribution of the training output $\boldsymbol{y}$ and test output $\boldsymbol{y}_{*}$ can be written explicitly as

\begin{equation}
\begin{bmatrix} \boldsymbol{y} \\ \boldsymbol{y}_{*} \end{bmatrix} \sim \mathcal{N} \left( \boldsymbol{0},  \begin{bmatrix*}[c] \boldsymbol{K} + \sigma_e^{2} \boldsymbol{I}_{n} & \boldsymbol{k}_{*} \\ \boldsymbol{k}_{*}^T & \boldsymbol{k}_{**} \end{bmatrix*} \right),
\end{equation}
where
\begin{itemize}
	\item $\boldsymbol{K} = \boldsymbol{K}(\boldsymbol{X}, \boldsymbol{X}; \boldsymbol{\theta})$ is an $n \times n$ kernel matrix of correlations among training inputs;
	\item $\boldsymbol{k}_{*} = \boldsymbol{K}(\boldsymbol{X}, \boldsymbol{X}_{*}; \boldsymbol{\theta})$ is an $n \times n_{*}$ kernel matrix of correlations between the training inputs and test inputs; $\boldsymbol{k}_{*}^T = \boldsymbol{K}(\boldsymbol{X}_{*}, \boldsymbol{X} ; \boldsymbol{\theta}) = \boldsymbol{K}(\boldsymbol{X}, \boldsymbol{X}_{*} ; \boldsymbol{\theta})^{T}$; 
	\item $\boldsymbol{k}_{**} = \boldsymbol{K}(\boldsymbol{X}_{*}, \boldsymbol{X}_{*}; \boldsymbol{\theta})$ is an $n_{*} \times n_{*}$ kernel matrix of correlations among test inputs.
\end{itemize}
By applying the results of conditional Gaussian distribution~\cite{RW06}, we can derive the posterior distribution as
\begin{equation}
p(\boldsymbol{y}_{*} \vert \mathcal{D}, \boldsymbol{X}_{*}; \boldsymbol{\theta}) \sim
\mathcal{N} \left(  \bar{\boldsymbol{\mu}} , \bar{\boldsymbol{\sigma}}  \right), 
\end{equation} 
where the posterior mean and variance are respectively given as
\begin{align}
\mathbb{E} \left[ f(\boldsymbol{X_*}) \right] &= \bar{\boldsymbol{\mu}} = \boldsymbol{k}_*^T \left( \boldsymbol{K} + \sigma_e^{2} \boldsymbol{I}_{n} \right)^{-1} \boldsymbol{y}, \label{eq:mean-func}\\
\mathbb{V} \left[ f(\boldsymbol{X_*}) \right]  &= \bar{\boldsymbol{\sigma}} = \boldsymbol{k}_{**} - \boldsymbol{k}_*^T \left( \boldsymbol{K} + \sigma_e^{2} \boldsymbol{I}_{n} \right)^{-1}  \boldsymbol{k}_*. \label{eq:var-func}
\end{align}
Note that the GP model is a Bayesian non-parametric model since the above posterior distribution can be refined as the number of observed data grows. 

\subsubsection{Kernel Function Tailored for Wireless Traffic Prediction}
\label{sec:kernel function}
Kernel function design is critical to GP, as it encodes the prior information about the underlying process.
A properly designed kernel function can generate both high modeling accuracy and strong generalization ability.
Generally, a kernel function can be either stationary or non-stationary. A \textit{stationary kernel} depends on the relative position of the two inputs, i.e., $ K(\boldsymbol{\tau})$ with $ \boldsymbol{\tau} = \boldsymbol{x_i} - \boldsymbol{x_j} $  and $ \boldsymbol{x_i}, \boldsymbol{x_j} $ being two different inputs. A \textit{non-stationary kernel} depends on the absolute position of the two inputs, i.e., $ K(\boldsymbol{x_i} , \boldsymbol{x_j})$. However, modeling with a stationary kernel is the preliminary requirement to generate a Toeplitz structure in the kernel matrix, which is the basis for the structured GP model proposed in Section \ref{sec:structured}. Therefore, we mainly focus on the stationary kernels in this paper.

Different stationary kernels can generate data profiles with distinctly different characteristics, e.g., a periodic kernel can generate structured periodic curves, while a squared exponential (SE) kernel can generate smooth and flexible curves. Moreover, multiple elementary kernels can be composed as a hybrid one while preserving their own particular characteristics. 
Additionally, in order to eliminate human interventions, automatic kernel learning/determination has become more and more fashionable in recent years. Representative results include~\cite{Gredilla10, WA13, Yin18} that proposed to design an universal kernel in the frequency domain and~\cite{DLG13} that proposed to search for a space of kernel structures built compositionally by adding and multiplying a small number of elementary kernels. The drawback of these advanced kernels lies in the increased model and training complexity.
In this paper, we prefer to use a relatively simpler linear kernel constructed based on our expert knowledge about the data generation process, not stretching to overfit the data for the robustness of a communication system. However, we note that automatic kernel learning is worth trying for better adaptivity and full automation of a system, especially when the training dataset is sufficiently large and the computation power is abundant. In the sequel, as a concrete solution, we select the following three kernels to model the wireless traffic patterns observed in Section~\ref{sec:patterns}.

Then, we compose them as a tailored kernel function particularly for wireless traffic prediction.
Specifically, 1) for the \textit{weekly periodic pattern}, we select a periodic kernel with the periodic length set to be the number of data points of one week, which is defined as
\begin{equation}\label{formula:periodickernel1}
k_1(t_i,t_j) = \sigma^2_{p_1} \exp\left[ -\frac{\sin^2{\left( \frac{\pi(t_i-t_j)}{\lambda_1} \right) }}{l_{p_1}^2} \right],
\end{equation}
where $ \lambda_1 $ is the periodic length, $ l_{p_1} $ is the length scale determining how rapidly the function varies with time $ t $, and $ \sigma^2_{p_1} $ is the variance determining the average distance of the function away from its mean;
2) for the \textit{daily periodic pattern}, we also select a periodic kernel, but with different hyper-parameters, which is defined as
\begin{equation}\label{formula:periodickernel2}
k_2(t_i,t_j) = \sigma^2_{p_2} \exp\left[ -\frac{\sin^2{\left( \frac{\pi(t_i-t_j)}{\lambda_2} \right) }}{l_{p_2}^2} \right],
\end{equation}
where the periodic length $ \lambda_2 $ is set to be the number of data points for one day, with length scale $ l_{p_2} $ and output variance $ \sigma^2_{p_2} $ also set differently from kernel $ k_1 $;
3) for the \textit{dynamic deviations}, we select an SE kernel, which is defined as
\begin{equation}\label{formula:SEkernel}
k_3(t_i,t_j) = \sigma^2_{l_t} \exp \left[ - \frac{(t_i-t_j)^2}{2 l^2_{l_t}} \right],
\end{equation}
where $ \sigma^2_{l_t} $ is the magnitude of the correlated components and $ l_t $ is its length scale.
The above three elementary kernels can be added together without influencing the GP properties~\cite{RW06}. Therefore, the tailored kernel function for general wireless traffic prediction tasks could be written as
\begin{equation}\label{key}
\begin{aligned}
k(t_i, t_j) = k_1(t_i,t_j)  + k_2(t_i,t_j) + k_3(t_i,t_j)
\end{aligned}
\end{equation}
with the kernel hyper-parameters 
\begin{equation}\label{key}
\bm{\theta}_h = \left[ \sigma_{p_1}^2, \sigma_{p_2}^2, \sigma_{l_t}^2,
l^2_{p_1}, l^2_{p_2}, l^2_{l_t}\right]^T. 
\end{equation}
Note that it is totally feasible to construct a new composite kernel function with other stationary kernels to predict traffic with different primary patterns. For example, the rational quadratic (RQ) kernel with three hyper-parameters is more flexible than the SE kernel to model the irregular dynamic deviations. Hence, the RQ kernel can be added into the composite kernel function to better model the traffic with obvious irregular variations, e.g., the future 5G traffic.
Meanwhile, it is noteworthy to point out that adding or deleting the \textit{stationary} elementary kernels will not influence the GP properties, such that our later proposed scalable GP framework still applies. 

\subsubsection{Learning Objectives}
The kernel function design determines the basic form of our GP regression model, where the predictive performance largely depends on the goodness of the model parameters, i.e., the hyper-parameters of kernel functions. 

Generally, the hyper-parameters can be initialized with a set of universal values when predicting different data profiles. However, specifying the initial hyper-parameters according to the observed primary patterns may help the hyper-parameter tuning start from a better initial point, thereby improving the learning efficiency. For example, when predicting similar curves to the one in the top panel of Fig.~\ref{fig:typical_traffic_week}, which shows a stronger weekly periodic pattern, the magnitude of the weekly periodic kernel $ \sigma^2_{p_1} $ could be initialized with a larger value than that of the other two kernels. After the initialization, the dominant method for tuning the model parameters is to maximize the marginal likelihood function, which can be written in a closed form as
\begin{equation}\label{eq:GPR-logMarginalFunction}
\log p(\boldsymbol{y}; \boldsymbol{\theta}) \! = \! -\!\frac{1}{2}\! \left(  \log |\boldsymbol{C}(\boldsymbol{\theta}) | \!+\! \boldsymbol{y}^T \boldsymbol{C}^{-1}(\boldsymbol{\theta}) \boldsymbol{y} \!+\! n \log(2 \pi) \right)\!,
\end{equation}
where $ |\cdot| $ is the matrix determinant and $\boldsymbol{C}(\boldsymbol{\theta}) \triangleq \boldsymbol{K}(\boldsymbol{\theta}_h) + \sigma_e^{2} \boldsymbol{I}_{n}$.
The model hyper-parameters $ \boldsymbol{\theta} $ can be tuned equivalently by minimizing the negative log-likelihood function. Therefore, the learning objective of our prediction model can be written as
\begin{equation}\label{P0}
\begin{aligned}
\mathcal{P}_{0}: \quad \min_{\boldsymbol{\theta}} \quad & l(\boldsymbol{\theta})  \!=\! \boldsymbol{y}^{T} \boldsymbol{C}^{-1}(\boldsymbol{\theta}) \boldsymbol{y} \!+\! \log  |\boldsymbol{C}(\boldsymbol{\theta})|, \\
\mathrm{s.t.} \quad & \boldsymbol{\theta} \in \Theta, \: \Theta \subseteq \mathbb{R}^p.
\end{aligned}
\end{equation}
It is noteworthy to remark that for most of kernel functions, no matter a standalone one or a composite one, problem $ \mathcal{P}_{0} $ is non-convex\footnote{However, for multiple linear kernels, such as the ones proposed in~\cite{6883125, add-ref1, Yin18}, $ \mathcal{P}_{0} $ becomes a difference-of-convex problem and efficient algorithms exist for solving the hyper-parameters. } and often shows no favorable structures in terms of the hyper-parameters. Consequently, the classic gradient descent methods, such as the limited-memory Broyden-Fletcher-Goldfarb-Shanno (L-BFGS) algorithm and the conjugate gradient methods could be used for solving the hyper-parameters but with no guarantee that the global minimum of $ \mathcal{P}_{0} $ could be found.
The work flow of the gradient descent based method is presented as follows. In each iteration, the hyper-parameters are updated as
\begin{equation}
\theta_{i}^{r+1} = \theta_{i}^{r} - \eta \cdot \frac{\partial l(\boldsymbol{\theta}) }{\partial \theta_i} \bigl\lvert_{\boldsymbol{\theta} = \boldsymbol{\theta}^{r}},  \quad \forall i = 1,2\cdots, p,
\end{equation}
where $\eta$ is the step size.  The derivative of the $i$-th element in $\boldsymbol{\theta}$ is computed as~\cite{RW06}
\begin{equation}\label{eq:gradient-step}
\frac{\partial l(\boldsymbol{\theta}) }{\partial \theta_i}\!\! = \Tr \left( \left( \boldsymbol{C}^{-1}(\boldsymbol{\theta}) - \boldsymbol{\gamma} \boldsymbol{\gamma}^{T} \right) \frac{\partial \boldsymbol{C}(\boldsymbol{\theta})}{\partial \theta_i} \right),
\end{equation}
where $ \Tr(\cdot) $ represents the matrix trace and $\boldsymbol{\gamma} \triangleq \boldsymbol{C}^{-1}(\boldsymbol{\theta}) \boldsymbol{y}$. 

Note that $\boldsymbol{C}^{-1}(\boldsymbol{\theta})$ has to be re-evaluated at each gradient step. Such a matrix inversion requires $\mathcal{O}(N^3)$ computations, where $ N $ is the number of training data, which dominates the computational complexity of the standard GP. Consequently, when the wireless traffic dataset is large, i.e., $ N $ is a large number, the time consumption of each execution grows exponentially, which prohibits its application from large-scale executions. Therefore, a scalable GP framework that can fully utilize the parallel computing resources in the BBU pool of C-RANs is desperately needed for large-scale applications.

\section{Scalable GP Framework with ADMM and Cross-Validation}
\label{sec:scalable gp}
\begin{figure*}
	\centering
	\subfigure[Training framework]
	{ 	
		\label{fig:mape}
		\includegraphics[trim=10 10 280 10, clip, height=5.5cm]{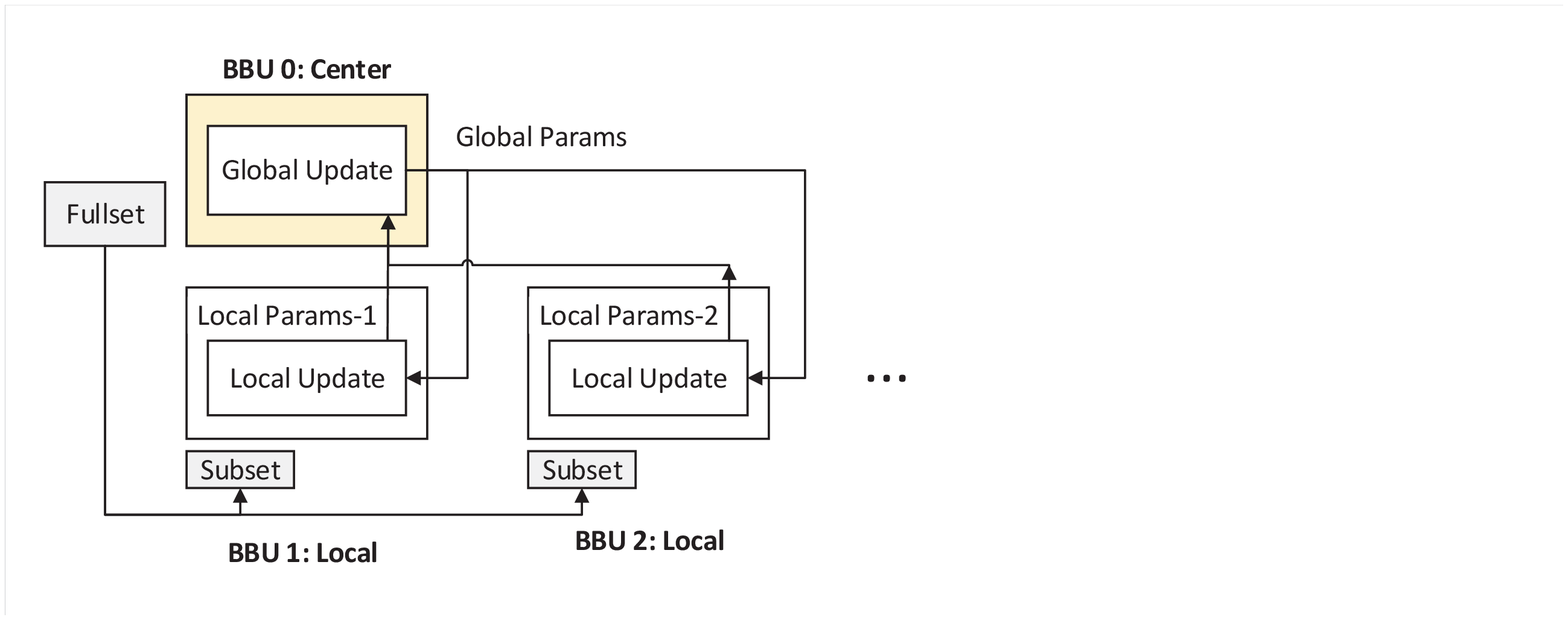}
	}
	\subfigure[Prediction framework]
	{ 	
		\label{fig:rmse}
		\includegraphics[trim=10 10 10 10, clip, height=5.5cm]{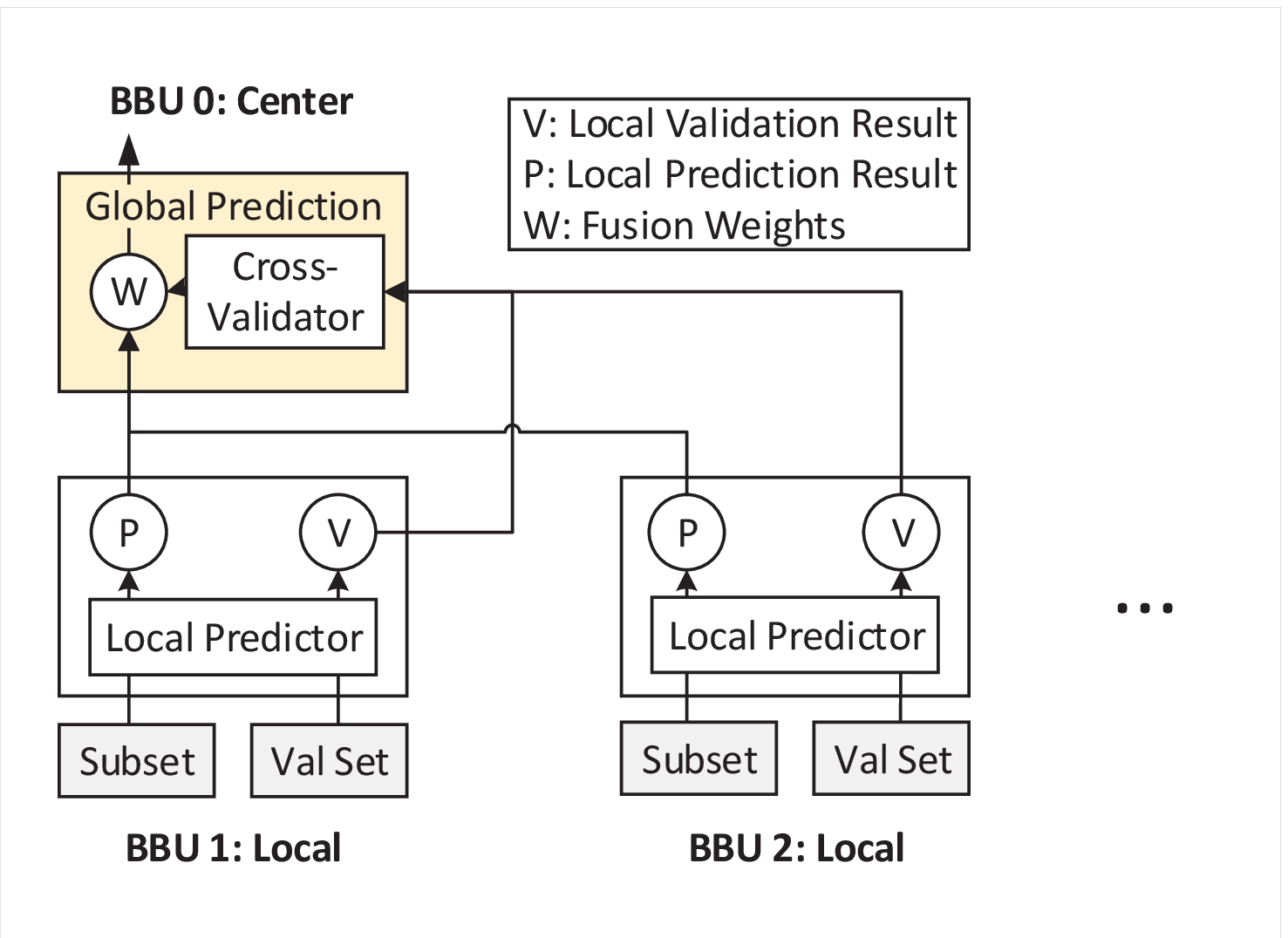}
	}
	\caption{Scalable GP framework with ADMM and Cross-Validation}
	\label{fig:framework}
\end{figure*}

In this section, we propose a scalable GP framework for the C-RAN based architecture given in Section \ref{sec:architecture}. The proposed framework can efficiently train the GP model via parallel computations based on multiple BBUs, and then optimally combine the local inferences from different BBUs to generate an improved global prediction. A detailed work-flow of the scalable GP framework is presented in Fig. \ref{fig:framework}.
Specifically, the framework contains one central node and multiple local nodes operate individually on parallel BBUs. 
At the training phase, each local node trains a GP model based on a subset of data split from the full training set and communicates with the central node for joint optimizations via the ADMM algorithm. 
At the prediction phase, each local node performs predictions on both validation points and test points based on its well-trained GP model. Afterwards, the central node first calculates the optimal fusion weights based on the prediction performance on validations points, then combines the local predictions on test points based on the calculated weights for a better and robust global prediction.

Generally, the central node is mainly responsible for 
i)~delivering the subsets of the full dataset to the non-central nodes at the very beginning; 
ii)~updating the global ADMM parameters based on the local ADMM parameters from other non-central nodes at the training phase; 
iii)~deriving the fusion weights based on the local predictions results from other non-central nodes at the prediction phase. 
Hence, it is feasible to transfer the functionalities of the central node to any of the non-central nodes by simply re-directing the data flow from other non-central nodes to the newly selected central node, and then, the global ADMM parameters and fusion weights can be fully recovered. 
On the other hand, malfunction of one non-central node will only cause the information loss of one subset, instead of a total failure.
However, it is noteworthy to point out that the centralized scheme can also be turned into a fully distributed scheme by letting each local node select a subset of data, update global parameters, and fuse weights, separately.
For example, 
i) each local node could select its own subset from the full dataset individually; 
ii) each local node could broadcast its local ADMM parameters, such that each of them is able to compute the global ADMM parameter individually; 
iii) each local node could broadcast its local prediction result, such that each of them is able to output the final prediction result individually. 
However, the information broadcast in the fully distributed scheme may increase the communication overhead in the C-RAN architecture.

With the proposed scalable GP framework, the computational complexity of training could be reduced from $ \mathcal{O}(N^3) $ of a standard GP to $ \mathcal{O}(\frac{N^3}{K^3}) $ of the proposed scalable GP, where $ N $ is the number of training points and $ K $ is the number of parallel BBUs, such that the dominating time consumption of GP regression could be largely decreased by simply increasing the number of BBUs. Moreover, the complexity could be further reduced to $ \mathcal{O}(\frac{N^2}{K^2}) $ when the kernel matrix follows a Toeplitz structure based on the structured GP model.
On the other hand, the computational complexity of fusion weights optimization scales as $ \mathcal{O}(\sqrt{\log K}) $, which makes the prediction phase also implementable for large-scale executions. 
As such, the dominant computational complexity of the central node scales as $ \mathcal{O}(\sqrt{\log K}) $, while the dominant computational complexity of each non-central node scales as $ \mathcal{O}(\frac{N^2}{K^2}) $, which makes the central node usually have a lower computational load than any of the non-central nodes.
	However, the central node does require a higher throughput than the non-central nodes to support the data delivery and collection, which should be considered in practical implementations.
In the following statements, we present the scalable GP training framework, scalable GP prediction framework, and structured GP in sequence.

\subsection{ADMM Empowered Scalable Training Framework}
\label{sec:ProposedGPADMM}
We aim to distribute the computation cost for solving $ \mathcal{P}_{0} $ evenly to a bunch of parallel BBUs for training speed-up.
Given the full training dataset $\mathcal{D} \triangleq \{\boldsymbol{X}, \boldsymbol{y} \}$, we define a set of $ K $ training subsets, denoted as $ \mathcal{S} \triangleq  \{\mathcal{D}^{(1)}, \mathcal{D}^{(2)}, \cdots, \mathcal{D}^{(K)} \} $. Each subset $ \mathcal{D}^{(i)} \triangleq  \{\boldsymbol{X}^{(i)}, \boldsymbol{y}^{(i)} \} $ is sampled from the full dataset $ \mathcal{D} $ and each local GP model is trained based on its own subset $ \mathcal{D}^{(i)}$. 

As in the PoE model~\cite{ng2014hierarchical}, we approximate the probability distribution of the full dataset by a product of the probability distributions of all local subsets, concretely,
\begin{subequations}
\begin{align}
p(\boldsymbol{y}|\boldsymbol{X}; \boldsymbol{\theta}) & \approx \prod_{i=1}^{K} p_i(\boldsymbol{y}^{(i)}|\boldsymbol{X}^{(i)}; \boldsymbol{\theta}), \label{eq:probability} \\
\log p(\boldsymbol{y}|\boldsymbol{X}; \boldsymbol{\theta}) & \approx \sum_{i=1}^{K} \log p(\boldsymbol{y}^{(i)}|\boldsymbol{X}^{(i)}; \boldsymbol{\theta}).
\end{align}
\end{subequations}
The standard parallel hyper-parameter training of the PoE can be expressed as
\begin{equation}\label{P1}
\begin{aligned}
\mathcal{P}_{1}: \quad \min_{\boldsymbol{\theta}} & \quad  \sum_{i=1}^{K} l^{(i)}(\boldsymbol{\theta}), \\
\mathrm{s.t.} & \quad \boldsymbol{\theta} \in \Theta, \\
\end{aligned}
\end{equation}
where
\begin{equation}\label{eq:local cost function}
l^{(i)}(\boldsymbol{\theta}) = (\boldsymbol{y}^{(i)})^{T} (\boldsymbol{C}^{(i)}(\boldsymbol{\theta}))^{-1} \boldsymbol{y}^{(i)} + \log  |\boldsymbol{C}^{(i)}(\boldsymbol{\theta})|,
\end{equation}
Hence, each local GP model only needs to optimize its own cost function as in Eq. (\ref{eq:local cost function}) \textit{w.r.t} the hyper-parameters. Such an optimization only requires operations on the small covariance matrix $ \boldsymbol{C}^{(i)} $, whose size is $ n_i \times n_i $, where $ n_i $ denotes the number of data points in the subset $ \mathcal{D}^{(i)}$ with $ n_i \ll n $. The computational complexity for each local GP model is thus reduced to $ \mathcal{O}(n_i^3) $ with standard GP implementation. Note that we use equal subset sizes for all local GP models in this paper, namely, $ n_i = \frac{N}{K}, \forall i = 1,2, \cdots, K $, which could be further optimized in our future work.

The philosophy behind PoE is to approximate the covariance matrix of full dataset with a block-diagonal matrix\footnote{If the local subsets are overlapped, the approximated matrix would not be a block-diagonal matrix~\cite{ng2014hierarchical}} of the same size. Each block is determined by the corresponding subset. Hence, the well-trained local hyper-parameters should ideally be the same, namely $ \boldsymbol{\theta_i} - \boldsymbol{\theta_j} = \boldsymbol{0}, \forall i,j $, to make the approximation from block-diagonal matrix consistent with the origin full matrix. 
Therefore, existing PoE-based methods~\cite{ng2014hierarchical, cao2014generalized, deisenroth2015distributed} assumed that all local GP models are trained jointly and shared the same set of global hyper-parameters $ \boldsymbol{\theta} $. However, such joint training can only be realized via rigorous gradient consensus. Specifically, for each gradient step in Eq. (\ref{eq:gradient-step}), the local cost and local derivative information should be collected and coordinated as the global cost and derivative. Such global information is then used to update a globally shared hyper-parameters $ \boldsymbol{\theta} $, which is transmitted back to each local GP model afterwards. 
However, the gradient-consensus steps, on the other hand, requires $ n_{\text{grads}} * \left( \dim(\boldsymbol{\theta}) + 1 \right) $ communication overheads for local cost and derivative collection, where $ n_{\text{grads}} $ is the number of gradient steps and $ \dim(\boldsymbol{\theta}) $ is the number of hyper-parameters to be optimized.
Meanwhile, the rigorous synchronous requirement per gradient step also restricts its practical application in real systems.

Therefore, we aim to empower the distributed GP model with the powerful ADMM algorithm to develop a principled parallel training framework. The ADMM-empowered framework allows each distributed units, e.g., the BBU in C-RANs, to be trained independently and to be coordinated with much less communication overhead. The proposed training framework could lay foundations for future scalable GP system design. 

In general, ADMM takes the form of a decomposition-coordination procedure, where the original large problem is decomposed into small local subproblems that can be solved in a coordinated way~\cite{Boyd11}. 
Based on ADMM, problem $ \mathcal{P}_{1} $ can be recast equivalently with the newly introduced local variables $ \boldsymbol{\theta}_i $ and a common global variable $\boldsymbol{z}$ as
\begin{equation}\label{P2}
\begin{aligned}
\mathcal{P}_{2}: \quad  \min_{\boldsymbol{\theta}_i}  \quad  & \sum_{i=1}^{K} l^{(i)}(\boldsymbol{\theta}_i), \\
\mathrm{s.t.}  \quad & \boldsymbol{\theta}_i - \boldsymbol{z} = \boldsymbol{0}, \quad i=1,2, \ldots, K, \\
\quad & \boldsymbol{\theta}_i \in \Theta, \quad i=1,2, \ldots, K. \\
\end{aligned}
\end{equation}
Note that the above two problems $ \mathcal{P}_1 $ and $ \mathcal{P}_2 $ are equivalent. But with the new formulation, each local GP model is free to train its own local hyper-parameters $ \boldsymbol{\theta}_i $ based on the local subset $ \mathcal{D}^{(i)}$, where the local hyper-parameters will eventually converge to the global hyper-parameter $\boldsymbol{z}$ after a few ADMM iterations, as presented below.

To solve problem $ \mathcal{P}_2 $ with ADMM, we formulate the augmented Lagrangian as
\begin{equation}
	\mathcal{L}^{(i)}(\boldsymbol{\theta}_i) \!\triangleq\! l^{(i)} (\boldsymbol{\theta}_i) \!+\! \sum_{i=1}^{K}  \boldsymbol{\zeta}^T_i (\boldsymbol{\theta}_i \!-\! \boldsymbol{z}) \!+\! \sum_{i=1}^{K} \frac{\rho}{2} \parallel \boldsymbol{\theta}_i \!-\! \boldsymbol{z} \parallel_2^2,
\end{equation}
where $ \boldsymbol{\zeta}_i $ is the dual variable and $ \rho > 0 $ is a fixed augmented Lagrangian parameter.
The sequential update of ADMM parameters in the $ (r+1) $-th iteration can be written as 
\begin{subequations}
	\begin{align}
	\boldsymbol{\theta}^{r+1}_i &:= \argmin_{\boldsymbol{\theta}_i} \mathcal{L}^{(i)}(\boldsymbol{\theta}_i, \boldsymbol{z}^r, \boldsymbol{\zeta}^{r}_i) \label{eq:local hyper-parameters}, \\
	\boldsymbol{z}^{r+1} &:= \frac{1}{K} \sum_{i=1}^{K} \left( \boldsymbol{\theta}^{r+1}_i + \frac{1}{\rho} \boldsymbol{\zeta}^{r}_i \right) \label{eq:global hyper-parameters}, \\
	\boldsymbol{\zeta}^{r+1}_i &:= \boldsymbol{\zeta}^{r}_i + \rho (\boldsymbol{\theta}^{r+1}_i - \boldsymbol{z}^{r+1}), \label{eq:dual variable}
	\end{align}
\end{subequations}
where the local $ \boldsymbol{\theta}_i $-minimization step in Eq. (\ref{eq:local hyper-parameters}) optimizes the local cost function and reduces the distance between the local and global estimates of the hyper-parameters at the same time.
The ADMM-based consensus only requires $ n_{\text{cons}} * \left( 2* \dim(\boldsymbol{\theta}) + 1 \right)  $ communication overheads, where $ n_{\text{cons}} $ is the number of ADMM iterations and we have $ n_{\text{cons}} \ll n_{\text{grads}}$.
Another benefit of such a ADMM-based consensus is that, if one local GP model is stuck at a bad local minimum, the global consensus may re-start at a more reasonable point for better converged result next time.

The optimality conditions for the ADMM solution are determined by the primal residuals $ \Delta_p $ and dual residuals $ \Delta_d $~\cite{Boyd11}. They can be given for each local GP model as
\begin{align}
\Delta_{i, p}^{r+1} &= \boldsymbol{\theta}_i^{r+1} - \boldsymbol{z}^{r+1}, i = 1,2,\cdots,K \label{eq:primal residuals},  \\
\Delta_d^{r+1 } &= \rho (\boldsymbol{z}^{r+1} - \boldsymbol{z}^{r} ) \label{eq:dual residuals}. 
\end{align}
The above two residuals will converge to zero as ADMM iterates. Hence, the stopping criteria for our problem comprises $ \norm{\Delta_{p}^{r}}_2 \leq \epsilon^{\text{pri}} $ and $ \norm{\Delta_d^{r}}_2 \leq \epsilon^{\text{dual}} $, where $ \epsilon^{\text{pri}} $ and $ \epsilon^{\text{dual}} $ are the feasibility tolerance constants for the primal and dual residuals, respectively. As suggested in~\cite{Boyd11}, they can be set as
\begin{align}
\epsilon^{\text{pri}} &= \sqrt{p} \, \epsilon^{\text{abs}} + \epsilon^{\text{rel}} \max \left\lbrace \norm{\boldsymbol{\theta}^r_i}_2, \norm{\boldsymbol{z}^r}_2 \right\rbrace, \label{eq:pri}\\
\epsilon^{\text{dual}} &= \sqrt{p} \, \epsilon^{\text{abs}} + \epsilon^{\text{rel}} \norm{\rho \boldsymbol{\zeta}^r}_2, \label{eq:dual}
\end{align}
where $ p $ denotes the dimension of $ \boldsymbol{\theta} $ in the $ l_2 $ norm.

The detailed ADMM-empowered GP training procedure is summarized in Algorithm \ref{algorithm:ADMM}. It should be pointed out that even for convex problems, the convergence of ADMM could be very slow~\cite{Boyd11}. But fortunately, a few iterations is often sufficient for ADMM to converge to an acceptable accuracy level in practical applications.
\begin{algorithm}[tb]
	\caption{ ADMM-empowered Scalable GP Training}
	\label{algorithm:ADMM}
	\begin{algorithmic}[1]
		\STATE \textbf{Initialization:} $ r=0 $, $ K $, $\boldsymbol{\theta}^0_i$, $\boldsymbol{\zeta}^0_i$, $ \rho $, $\boldsymbol{z}^0 = \frac{1}{K} \sum_{i=1}^{K} \left( \boldsymbol{\theta}^0_i + \frac{1}{\rho} \boldsymbol{\zeta}^{0}_i \right) $, tolerance $\epsilon^{\text{abs}}$, $\epsilon^{\text{rel}}$.\\
		\STATE \textbf{Iteration:}
		\WHILE{$ || \Delta_{i, p}^{r} ||^2 \geq \epsilon^{\text{pri}} $ or $ ||\Delta_d^{r}||^2 \geq \epsilon^{\text{dual}} $ }
		\STATE{$ r = r+1 .$}
		\FOR{$ i = 1, \cdots, K $}
		\STATE{Obtain the parameters for local GP $ i $ by Eq. (\ref{eq:local hyper-parameters}).}
		\ENDFOR
		\STATE{Obtain the global parameters by Eq. (\ref{eq:global hyper-parameters}). }
		\STATE{Obtain the dual variable by Eq. (\ref{eq:dual variable}). }
		\STATE {Calculate the primal residuals $ \Delta_{i, p}^{r+1} $ and the dual residuals $ \Delta^{r+1}_d $ by Eq. (\ref{eq:primal residuals}) and Eq. (\ref{eq:dual residuals}), respectively. }
		\STATE Update the feasibility tolerance $ \epsilon^{\text{pri}} $ and $ \epsilon^{\text{dual}} $ by Eq. (\ref{eq:pri}) and Eq. (\ref{eq:dual}).
		\ENDWHILE\\
		\STATE \textbf{Output:} Global parameter $ \boldsymbol{z} $.
	\end{algorithmic}
\end{algorithm}

\subsection{Cross-Validation based Scalable Fusion Framework}
\label{sec:prediction phase}
After having trained the GP model hyper-parameters, we need to fuse the local predictions from all BBUs to get a global prediction. In contrast to the existing fusion strategies that are based on empirical weights, e.g., using the entropy as weights directly in~\cite{cao2014generalized, deisenroth2015distributed}, we propose to optimize the weights via cross-validation, which could provide a reliable fusion quality with concrete theoretical analysis. Meanwhile, as suggested in~\cite{cao2014generalized}, we aim to achieve three desirable properties for the fusion process: 1) the predictions are combined based on the prior or posterior information rather than being fixed, which gives the combined model more generalization power; 2) the combination should follow a valid probabilistic model, which helps preserve the distinct GP properties, e.g., the posterior variance for prediction uncertainty evaluation; 3) the combined prediction should alleviate the influence from bad local predictions, which makes the global prediction robust against bad local minimals. 


In the following statement, we first present the generalized PoE framework discussed in~\cite{cao2014generalized, deisenroth2015distributed}, which can assign explicit weighting terms to local predictions while preserving the probabilistic GP structure. Then we propose our cross-validation based weighting model, and show that the optimization problem could be expressed in a convex form to obtain the global optimal solution under certain conditions. In other cases, the optimization problem can be solved via the mirror descent method for locally optimal solution efficiently with convergence guarantees.
Finally, we present a simplified weighting strategy that does not require information sharing and has a constant computational cost.

The generalized PoE model proposes to bring in an explicit weight parameter $ \beta $ in the prediction phase, to balance the importance among different local predictions. The revised predictive distribution is
\begin{equation}\label{eq:gpoe-predictive}
p(f_*|\boldsymbol{x}_*, \mathcal{D}) \approx \prod_{i=1}^{K} p_i^{\beta_i}(f_*|\boldsymbol{x}_*, \mathcal{D}^{(i)}),
\end{equation}
where $ \beta_i $ is the weight for the $i$-th local GP model and the corresponding posterior mean and variance are, respectively,
\begin{align}
	\mu_*    &= (\sigma_*)^2 \sum_{i=1}^{K} \beta_i \sigma_i^{-2}(\boldsymbol{x}_*)\mu_k(\boldsymbol{x}_*), \label{eq:gmean}\\
	\sigma^2_* &= \left( \sum_{i=1}^{K} \beta_i \sigma_i^{-2}(\boldsymbol{x}_*) \right)^{-1} \label{eq:gvar}.
\end{align}
Consequently, the choice of $ \beta $ is vital to the prediction phase. Existing works, e.g.,~\cite{cao2014generalized, deisenroth2015distributed}, exploited the differential entropy as the weights. However, the entropy-based weights cannot guarantee a robust improvement on the prediction accuracy. Besides,~\cite{cao2014generalized} also pointed out that the lack of change in entropy did not necessarily mean an irrelevant prediction: e.g., the kernel could be mis-specified, which indicates that the differential entropy may not always be reliable.

\begin{figure}
	\centering
	\includegraphics[trim = 20 10 20 10, clip, width = 7cm]{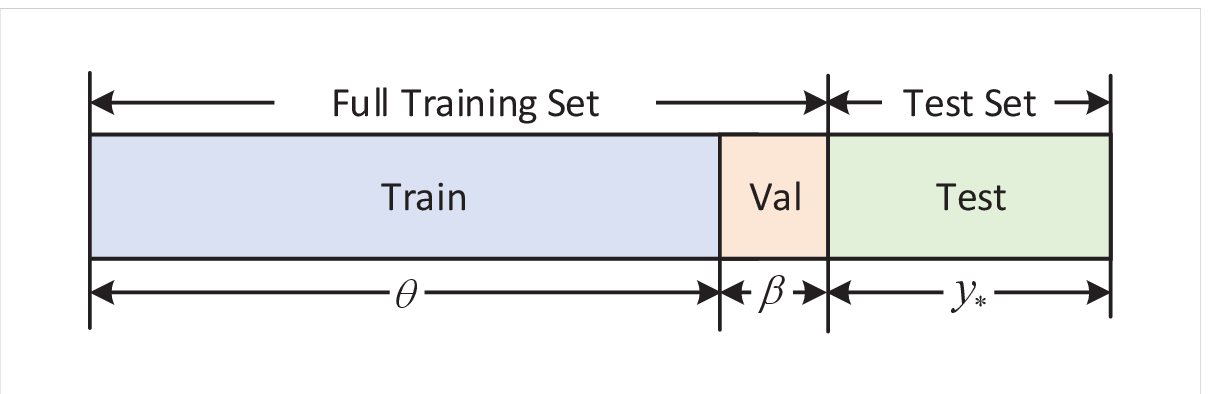}
	\caption{The full training set for each RRH is split into a training set and a validation set, where the former is used in the hyper-parameters $ \theta $ learning while the latter is used in the fusion weights $ \beta $ optimization. The test set remains the same, which is the prediction target $ y_* $. }
	\label{fig:division}
\end{figure}
For general regression tasks, the points with shorter input distances are usually deemed, with a higher probability, as having similar target values, implying that training points that are closer to the desired test points should be more informative when performing predictions~\cite{RW06}.
Hence, as shown in Fig. \ref{fig:division}, we divide the full training set into two parts, i.e., the training set and the validation set, where the validation set consists of the training points that are closer to the test set. 
In order to guarantee a robust performance for general regression tasks, we first optimize the prediction performance on the validation set, then use the optimized weights to combine local inferences for test set. 
Specifically, optimizing the prediction performance on the validation set can be formulated as the minimization of the prediction residuals:
\begin{equation}\label{opt-pred-origin}
\begin{aligned}
\min_{\bm{\beta}} \quad & \sum_{m=1}^{M} \left( y_m - \tilde{y}_m  \right)^2 , \\
\mathrm{s.t.}  \quad & \boldsymbol{\beta} \in \Omega,
\end{aligned}
\end{equation}
where $ M $ is the size of the validation set, $ \tilde{y}_m $ is the combined prediction on the validation point $ m $ and $ \Omega = \left\lbrace  \bm{\beta} \in \mathbb{R}^K_{+}: \bm{e}^T\bm{\beta} = 1 \right\rbrace $ restricts the weights $ \boldsymbol{\beta} $ to be in a probability simplex. 

Based on the joint posterior estimation from parallel BBUs given by Eq. (\ref{eq:gpoe-predictive}) - (\ref{eq:gvar}), the combined prediction $ \tilde{y}_m $ can be written as
\begin{subequations}\label{key}
	\begin{align}
	\tilde{y}_m = \: & \argmax_{\tilde{f}_m} \prod_{k=1}^{K} p^{\beta_{k}} \left( \tilde{f}_m|\mu_k(\boldsymbol{x}_m), \sigma_k(\boldsymbol{x}_m)\right)\\
	= \: & \sigma_m^2 \sum_{i=1}^{K} \beta_i \sigma_i^{-2}(\boldsymbol{x}_m)\mu_k(\boldsymbol{x}_m) \\
	= \: & \frac{\sum_{i=1}^{K} \beta_i \sigma_i^{-2}(\boldsymbol{x}_m)\mu_k(\boldsymbol{x}_m)}{\sum_{i=1}^{K} \beta_i \sigma_i^{-2}(\boldsymbol{x}_m)}.
	\end{align}
\end{subequations}
Therefore, the optimization problem proposed in Eq. (\ref{opt-pred-origin}) can be re-cast as
\begin{equation}\label{P3}
\begin{aligned}
\mathcal{P}_{3}: \quad \min_{\bm{\beta}} \quad & f(\bm{\beta}) = \sum_{m=1}^{M} \left( y_m - \frac{\sum_{i=1}^{K} a_i(x_m) \beta_i }{\sum_{i=1}^{K} b_i(x_m) \beta_i } \right)^2 , \\
\mathrm{s.t.} \quad & \boldsymbol{\beta} \in \Omega,
\end{aligned}
\end{equation}
where
\begin{subequations}\label{key}
\begin{align}
a_i(x_m) &= \sigma_i^{-2}(x_m) \mu_i(x_m),  \\
b_i(x_m) &= \sigma_i^{-2}(x_m). 
\end{align}
\end{subequations}
The convexity of problem $ \mathcal{P}_{3} $ depends on the size of the validation set. Specifically, when we use a single point for validation, i.e., $ M =1 $, problem $ \mathcal{P}_{3} $ can be cast into a convex form, where the global optimal point can be obtained; when we use more than one points for validation, i.e., $ M > 1 $, problem $ \mathcal{P}_{3} $ will be non-convex, but can be solved rather efficiently for suboptimal solutions. Detailed analysis is conducted as follows.

\subsubsection{Global optimal solution with a single validation point}
When using a single point for validation, the objective function in problem $ \mathcal{P}_{3} $ could be simplified as
\begin{subequations}\label{key}
\begin{align}
	f(\bm{\beta}) = & \left( y_* - \frac{\sum_{i=1}^{K} a_i(x_*) \beta_i }{\sum_{i=1}^{K} b_i(x_*) \beta_i } \right)^2 \\
	= & \left( y_* - \sum_{i=1}^{K} a_i(x_*) r_i \right)^2,
\end{align}
\end{subequations}
where $ r_k = \beta_k / \sum_{i=1}^{K} b_i(x_*) \beta_i $, such that we can transform problem $ \mathcal{P}_{3} $ into a classic quadratic programming (QP) problem as
\begin{equation}\label{P4}
\begin{aligned}
\mathcal{P}_{4}: \quad \min_{\bm{r}} \quad & f(\bm{r}) = \left( y_* - \sum_{i=1}^{K} a_i(x_*) r_i \right)^2 , \\
\mathrm{s.t.}  \quad & \sum_{i=1}^{K} b_i(x_*)r_i = 1, \\
\quad & r_i \geq 0, i = 1,2,\cdots, K.
\end{aligned}
\end{equation}
The global optimal solution $ \boldsymbol{r}^* $ for problem $ \mathcal{P}_{4} $ can be easily obtained. Finally, the optimal weights can be calculated as $ \beta_i^* = \frac{r_i^*}{\sum_i r_i^*} $.

\subsubsection{Local optimal solution with multiple validation points}
When using multiple points for validation, we could use mirror descent to solve problem $ \mathcal{P}_{3} $ for locally optimal solutions. The mirror descent method is a first-order optimization procedure which provides an important generalization of the sub-gradient descent method towards non-Euclidean geometries~\cite{beck2003mirror}, and has been applied to many large-scale optimization problems in machine learning applications, e.g., online learning~\cite{srebro2011universality}, multi-kernel learning~\cite{jagarlapudi2009algorithmics}. The main motivation for applying mirror descent to problem $ \mathcal{P}_{3} $ is that, mirror descent can outperform the regular projected sub-gradient method when dealing with optimization problems on high-dimensional spaces~\cite{duchi2010composite, beck2003mirror}, which in our case, refers to the number of BBUs. 

We first approximate the prediction residual $ f(\bm{\beta}) $ around $ \bm{\beta}^r $ by the first-order Taylor expansion as
\begin{equation}\label{key}
f(\bm{\beta}) \approx f(\bm{\beta}^r) + \langle g(\bm{\beta}^{r}), \bm{\beta} - \bm{\beta}^{r} \rangle.
\end{equation}
Then, we penalize such displacement with a Bregman divergence term, which makes the update of $ \boldsymbol{\beta} $ as
\begin{equation} 
\label{eq:argminbeta}
\bm{\beta}^{r+1} = \argmin_{\bm{\beta} \in \Omega} q(\bm{\beta}^r), 
\end{equation}
where
\begin{equation}\label{key}
q(\bm{\beta}^r) \!=\! f(\bm{\beta}^{r}) \!+\! \langle g(\bm{\beta}^{r}), \bm{\beta} \!-\! \bm{\beta}^{r} \rangle \!+\! \frac{1}{\eta^{r}} D_{\psi}(\bm{\beta}, \bm{\beta}^{r}),
\end{equation}  
with $ g(\bm{\beta}^{r}) $ the gradient of $ f(\bm{\beta}^{r}) $ and $ D_{\psi}(\bm{\beta}, \bm{\beta}^{r}) = \psi(\bm{\beta})-\psi(\bm{\beta}^{r}) -\nabla\psi(\bm{\beta}^{r})^T(\bm{\beta}-\bm{\beta}^{r}) $
the Bregman divergence.
Consistent with the sub-gradient descent rule, the update of Eq. (\ref{eq:argminbeta}) can be resolved as
\begin{subequations}
	\begin{align}
	\bm{\beta}^{r+\frac{1}{2}} &= \argmin_{\bm{\beta}} q(\bm{\beta}^r), \label{eq:1}\\
	\bm{\beta}^{r+1} &= \argmin_{\bm{\beta} \in \Omega} D_{\psi}(\bm{\beta}, \bm{\beta}^{r+\frac{1}{2}}). \label{eq:2}
	\end{align}
\end{subequations}

We use the Kullback-Leibler (KL) divergence for $ D_{\psi}(\bm{\beta}, \bm{\beta}^{r}) $ in our problem, where $ \psi(x) = \sum_{i=1}^{K} x_i \log x_i$ with $i = 1,2,\cdots, K$.
The subproblem given by Eq. (\ref{eq:1}) is a convex problem. By taking the gradient with respect to $ \bm{\beta} $, we obtain the optimal condition of a specific local GP model $ i $ as
\begin{subequations}
	\begin{align}
	&  g(\beta_i^{r}) + \frac{1}{\eta^r} \left( \nabla \psi (\beta_i^{r+\frac{1}{2}}) - \nabla \psi (\beta_i^{r}) \right) = 0, \\
	\Longleftrightarrow \quad & \nabla \psi (\beta_i^{r+\frac{1}{2}}) = \nabla \psi (\beta_i^{r}) - \eta^r g(\beta_i^{r}), \\
	\Longleftrightarrow \quad & \log (\beta_i^{r+\frac{1}{2}}) = \log (\beta_i^{r+\frac{1}{2}}) - \eta^r g(\beta_i^{r}), \\
	\Longleftrightarrow \quad & \beta^{r+\frac{1}{2}}_i = \beta^r_i \exp \left\lbrace -\eta^r g_i \right\rbrace.
	\end{align}
\end{subequations}
The subproblem given by Eq. (\ref{eq:2}) over the simplex $ \Omega = \left\lbrace  \bm{\beta} \in \mathbb{R}^K_{+}: \bm{e}^T\bm{\beta} = 1 \right\rbrace $ is also a convex problem, where the Lagrangian function can be written as
\begin{equation}\label{key}
\mathcal{L} \!=\! \sum^K_{i=1} \beta_i \log \frac{\beta_i}{\beta^{r+\frac{1}{2}}_i} \!-\! \sum^K_{i=1} \left( \beta_i \!-\! \beta^{r+\frac{1}{2}}_i\right)  \!+\! \lambda \left( \sum_{i=1}^{K}\beta_i \!-\! 1 \right).
\end{equation}
For all $ i = 1, \cdots , K $, we have
\begin{equation}\label{key}
\frac{\partial \mathcal{L}}{\partial \beta_i} = \log \frac{\beta_i}{\beta^{r+\frac{1}{2}}_i} + \lambda = 0.
\end{equation}
Thus, we have $ \beta_i = \gamma \beta^{r+\frac{1}{2}}_i, \forall i = 1,2, \cdots, K$. Given $ \bm{e}^T\bm{\beta} = 1 $, we have $ \gamma = \frac{1}{\bm{e}^T \bm{\beta}^{r+\frac{1}{2}}} $. Therefore, the projection with respect to the KL divergence in Eq. (\ref{eq:2}) amounts to a simple renormalization as
\begin{equation}\label{key}
\beta^{r+1}_i = \frac{\beta^{r+\frac{1}{2}}_i}{\bm{e}^T \bm{\beta}^{r+\frac{1}{2}}} .
\end{equation}
Therefore, the update rules given in Eq. (\ref{eq:1}) and Eq. (\ref{eq:2}) become, respectively,
\begin{subequations}\label{eq:mirror-descent}
	\begin{align}
	\beta^{r+\frac{1}{2}}_i &= \beta^r_i \exp \left\lbrace -\eta^r g_i^r \right\rbrace, \label{eq:mirror-1} \\
	\beta^{r+1}_i &= \frac{\beta^{r+\frac{1}{2}}_i}{\bm{e}^T \bm{\beta}^{r+\frac{1}{2}}}. \label{eq:mirror-2}
	\end{align}
\end{subequations}
The convergence analysis is presented as follows. Bounding by  $ ||g^r_i||^2_2 \leq G $ and $||D_{\psi}(\bm{\beta}^*, \bm{\beta}^{1})|| \leq R, \forall i = 1, \cdots , K $, the convergence rate of mirror descent with KL divergence in Eq. (\ref{eq:mirror-descent}) can be given by~\cite{beck2003mirror}
\begin{equation}\label{eq:bound}
\min_{ 1 \leq r \leq T } \epsilon_k \leq \frac{2R^2 + G^2\sum_{r=1}^T (\eta^r)^2 }{2 \sum_{r=1}^T \eta^r},
\end{equation}
where $ \epsilon_k = f(\boldsymbol{\beta^r}) - f(\boldsymbol{\beta^*}) $ and $ \boldsymbol{\beta^*} $ is the global optimum point. The upper bound given in Eq. (\ref{eq:bound}) is a convex and symmetric function of $ \eta^r $, such that the optimal upper bound can be achieved when setting a constant step-size, i.e., $ \eta^r = \frac{R}{G} \sqrt{\frac{2}{T}}  $,  which reduces the upper bound to be
\begin{equation}\label{key}
\min_{ 1 \leq r \leq T } \epsilon_k \leq RG \sqrt{\frac{2}{T}},
\end{equation}
where $ T $ denotes the number of iterations.
With $ \beta^1_i = n^{-1} $ and $ \sum_{i=1}^{K} \beta_i^* = 1, \beta_i^* >0 $, we have $ R = \sqrt{\log K} $.  
The detailed procedures are presented in Algorithm \ref{algorithm:mirror}.

\begin{algorithm}[tb]
\caption{Mirror Descent for Weight Optimization}
\label{algorithm:mirror}
\begin{algorithmic}[1]
	\STATE \textbf{Initialization:} $r=0$, $ \boldsymbol{\beta}^0 $, $\eta^r = \eta^0$, tolerance $ \epsilon^{\text{mirror}} $.
	\STATE \textbf{Iteration:}
	\WHILE{$|f(\boldsymbol{\beta^r}) - f(\boldsymbol{\beta^*})| \geq \epsilon^{\text{mirror}} $}
	\STATE $r=r+1$.
	\FOR{$ i = 1, \cdots, K $}
	\STATE Optimize the unconstrained weights by Eq. (\ref{eq:mirror-1}).
	\STATE Re-normalize the weights by Eq. (\ref{eq:mirror-2}).
	\ENDFOR
	\ENDWHILE\\
	\STATE \textbf{Output:} Optimal fusion weights $\boldsymbol{\beta}$.
\end{algorithmic}
\end{algorithm}

\subsubsection{Simplified soft-max based solution}
Intuitively, the validation dataset should be carefully selected to mimic the test data profile so as to achieve overally better prediction performance. The computational complexity of mirror descent based weight optimization increases along with the number of BBUs, which may not be able to meet the response time requirement when using massive parallel BBUs for delay-sensitive applications. 
Hence, we propose a simplified fusion weight optimization method based on the soft-max function. Specifically, we set the weights to be inversely proportional to the average prediction error on the validation set, and we smooth such proportion with a soft-max function as
\begin{equation}\label{key}
\beta_k = \frac{\exp(-e_k)}{\sum_{k=1}^{K} \exp(-e_k)},
\end{equation}
where $ e_k $ is the averaged error (e.g., the root mean square error (RMSE)) obtained for the $k$-th local GP model on the validation dataset. Note that the soft-max based weighting strategy does not involve any joint optimization and has a constant computational complexity. But the fusion performance is slightly degraded compared with the previous optimization-based methods. Detailed comparisons are presented in the experimental results.


\subsection{Extra Complexity Reduction for Structured Kernel Matrices}
\label{sec:structured}
As pointed out in Section \ref{sec:system model}, GP regression models need to take the inverse and log-determinant of a kernel matrix at each iteration of the classic gradient descent method when learning the optimal hyper-parameters. The standard implementation of Eq. (\ref{eq:gradient-step}) relies on the Cholesky factorization~\cite{RW06}, which is computationally demanding when dealing with a large dataset. However, when there is a certain favorable structure in the kernel matrix, the computational complexity can be further reduced.
For instance, the regularly recorded wireless traffic data from RRHs form a regularly-spaced time series, which is common for the standard wireless systems.
Additionally, the tailored kernel function for wireless traffic prediction proposed in Section \ref{sec:kernel function} is composed of stationary kernels. These will give rise to a Toeplitz structure in the kernel matrix, such that fast matrix operations can be applied to reduce the computational complexity. Consequently, each BBU could exploit the structured GP method to generate a reduced $\mathcal{O}(\frac{N^2}{K^2})$ computational complexity from aforementioned $\mathcal{O}(\frac{N^3}{K^3})$ for further speed acceleration.
More details about the structured GP can be found in our previous work~\cite{8254808}. 

\section{Experimental Results}
\label{sec:results}
\subsection{Data Description and Pre-processing}
The experimental dataset contains hourly recorded downlink PRB usage histories of 3072 base stations in three southern cities of China, from September 1st to September 30th, 2015. Each traffic flow history profile on each day contains 24 data points corresponding to 24 hours.
\begin{figure}
	\centering
	\includegraphics[trim = 20 10 20 20, clip, width = 8.5cm]{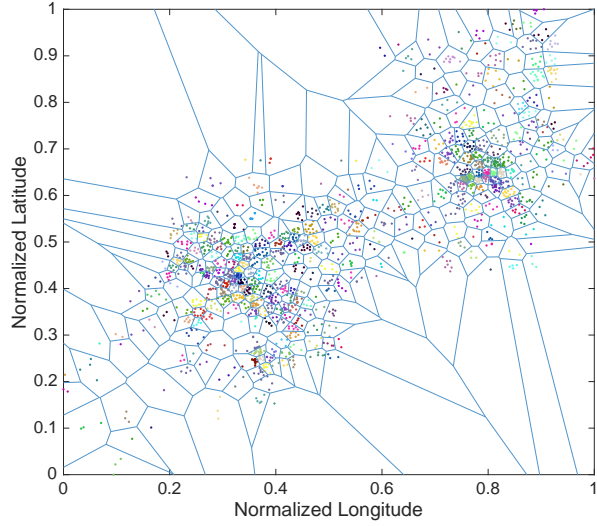}
	\caption{Each polygon represents an area served by a certain RRH cluster.}
	\label{fig:cluster}
\end{figure}
The base stations can be treated as the RRHs in the C-RANs.
As the 4G networks have only been commercially used for less than one year when the data was collected, each cell usually only contains less than ten active 4G users per hour. To set a more realistic data profile, we group the RRHs (the cells) into 360 clusters according to their geographical distribution by the K-means clustering algorithm, and the result is shown in Fig. \ref{fig:cluster}. Accordingly, each RRH cluster provides services for a certain area with the averaged coverage range around 1km.
The PRB usages are aggregated among all RRHs within the same cluster to obtain the wireless traffic of its served area, which is our prediction target. The number of users in each RRH cluster per hour is thus increased by a rough order of one hundred.
Consequently, as shown in Fig.~\ref{fig:typical_traffic_week}, the aggregated traffic exhibits obvious periodic traffic patterns along with strong dynamic deviations that are highly likely to appear in the 5G scenarios, as discussed in Sec.~\ref{sec:patterns}. 

\subsection{Performance Metrics}
In this paper, we use the RMSE and the mean absolute percentage error (MAPE) averaged over multiple test points as the performance metric:
\begin{subequations}
\begin{align}
	e_{\text{RMSE}} &= \sqrt{ \frac{\sum_{i=1}^{n_*}\left( y_* (\boldsymbol{x}_i) - y(\boldsymbol{x}_i) \right)^2}{n_*} }, \\
	e_{\text{MAPE}} &=  \frac{1}{n_*} \sum_{i=1}^{n_*} \left| \frac{ y_* (\boldsymbol{x}_i) - y(\boldsymbol{x}_i) }{y(\boldsymbol{x}_i)} \right|  \times 100, %
\end{align}
\end{subequations}
where $ n_* $ is the number of test data points, $ y_* (\boldsymbol{x}_i) $ is the posterior GP prediction for the test input $ \boldsymbol{x}_i $, and $ y(\boldsymbol{x}_i) $ is the ground truth.

\subsection{Results Analysis}
In the following experiments, we consider three baselines to present the performance comparison between the GP and other prediction models: the seasonal ARIMA model~\cite{shu2003wireless}, the sinusoid superposition model~\cite{wang2015approach}, and the recurrent neural network with long short-term memory (LSTM) based on deep learning techniques~\cite{8264694}. We refer to the baseline models as SARIMA, SS, and LSTM respectively for short.
Meanwhile, we consider another three baselines to present the performance comparison between our proposed scalable GP framework and other scalable GP models: the full GP, the rBCM model based on the distributed GP~\cite{deisenroth2015distributed}, and the subset-of-data (SOD) model based on the sparse GP~\cite{Chalupka}. The full GP operates on the full dataset. The rBCM operates on subsets, generated by random data assignments, and the training for rBCM operates in the gradient-consensus manner as described in Section \ref{sec:ProposedGPADMM} to generate its best performance. The SOD uses a random subset of the full training data as the training set, which was identified as a good and robust sparse GP approximation in~\cite{deisenroth2015distributed, Chalupka}.


\begin{table}[t]
	\centering
	\caption{Time Consumption for Training Phase}
	\label{table:training}
	\begin{tabular}{@{}cccccc@{}}
		\toprule
		\textbf{Model} & \textbf{1 BBU} & \textbf{2 BBUs} & \textbf{4 BBUs} & \textbf{8 BBUs} & \textbf{16 BBUs} \\ \midrule
		STD            & 16.8s          & 3.5s            & 1.1s            & 0.4s            & 0.1s             \\ \midrule
		TPLZ           & 6.9s           & 1.2s            & 0.4s            & 0.2s            & 0.1s             \\ \midrule
		rBCM~\cite{deisenroth2015distributed}           & 16.8s          & 4.9s           & 2.4s            & 0.4s            & 0.2s             \\ \bottomrule
	\end{tabular}
\end{table}

\begin{table}[t]
	\centering
	\caption{Time Consumption for Prediction Phase}
	\label{table:prediction}
	\begin{tabular}{@{}ccccc@{}}
		\toprule
		\textbf{Weight Model} & \textbf{2 BBUs} & \textbf{4 BBUs} & \textbf{8 BBUs} & \textbf{16 BBUs} \\ \midrule
		Mirror & 0.07s & 0.13s & 0.21s & 0.37s \\ \midrule
		Soft-max & 0.06s & 0.05s & 0.03s & 0.03s \\ \midrule
		rBCM~\cite{deisenroth2015distributed} & 0.08s & 0.06s & 0.06s & 0.05s \\ \bottomrule
	\end{tabular}
\end{table}

\subsubsection{Operation Time}
In Table~\ref{table:training}, we show that per each execution, the training time of our scalable training framework can be largely reduced by increasing the number of running BBUs. Moreover, the structured GP (TPLZ) can further reduce the time consumption of the standard GP (STD). In Table~\ref{table:prediction}, we show that the prediction time is much less than the training time, especially for the soft-max based weight model. 
The simulations are performed on our 4G wireless traffic dataset from one RRH cluster with 700 data points.
The BBU pool is simulated based on a workstation with eight Intel Xeon E3 CPU cores at 3.50 GHz and 32 gigabytes memory. The number of BBUs varies from two to sixteen with the same computing power, which are simulated by activating them on the workstation in rotation.
In Table~\ref{table:training}, each BBU trains a local GP model based on a subset with the size of $ \frac{N}{K}$, sampled from the full dataset, where $ K $ is the number of BBUs and $ N $ is the size of the full dataset, i.e., 700. To guarantee a fair comparison, we initialize all local GP models with the same hyper-parameters in all experiments, and average the results over 100 runs.
%
In Table~\ref{table:prediction}, we use three validation points in the weight optimization for all simulations. The result also indicates that when using the mirror descent based fusion method, the time consumption increases slowly with increased number of BBUs. However, for the soft-max based fusion method, the time consumption decreases slowly, for that the fusion weight optimization no longer dominates the time consumption, instead, the posterior inference takes place, which will become faster when being executed with smaller subsets.

As for the time consumption of the comparing schemes, i)~as shown in Table~\ref{table:training} and Table~\ref{table:prediction}, the gradient-consensus based training phase of the rBCM exhibits a similar time consumption to that of our scalable training framework with STD (without the acceleration from the structured GP). The entropy-weighting based prediction phase of rBCM exhibits a similar time consumption to our soft-max based prediction phase with cross-validation; 
ii)~the SARIMA, SS, and LSTM methods require about $ 4 $s, $ 2 $s, and $ 270 $s per single execution, respectively, using the full computing power of the same workstation. 
Noticeably, with the acceleration from the structured GP, our proposed scalable GP framework can run faster than the SARIMA, SS, and LSTM models when using more than two BBUs for parallel computation. Moreover, in what follows, we will show that the prediction performance of our scheme can also surpass the competing schemes considerably. 

%

\subsubsection{Performance of Full GP}
In the following experiments, we present the performance comparisons between the GP and other prediction models. For all baseline models, we train them with the full dataset, and consider their predictions based on temporal information only, consistent with our GP model. We use one BBU to train our proposed GP-based model directly on the full dataset to generate the best prediction performance, which is referred to as \textit{full GP} for short. Note that the training and prediction phases of the full GP model are consistent with the standard GP model, which is centralized without the scalable framework. 

\begin{figure*}
	\centering
	\subfigure[One-hour look-ahead. ]
	{ 	
		\centering
		\label{fig:prediction}
		\includegraphics[trim=60 30 60 20, clip, height=6cm]{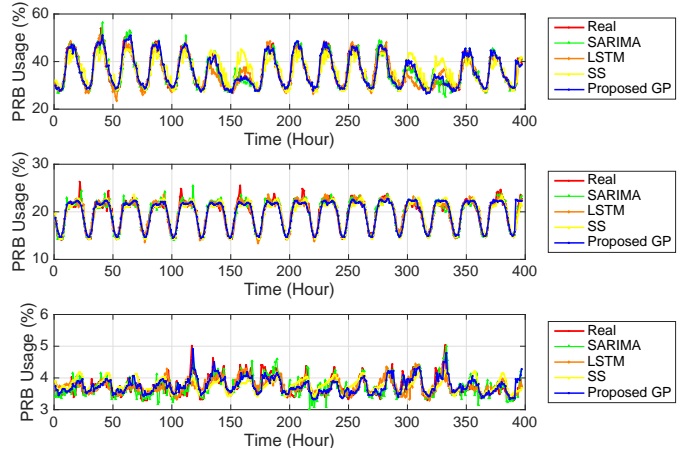}
	}
	\subfigure[Twenty-four-hour look-ahead]
	{ 	
		\centering
		\label{fig:var}
		\includegraphics[trim=30 25 50 20, clip, height=6cm]{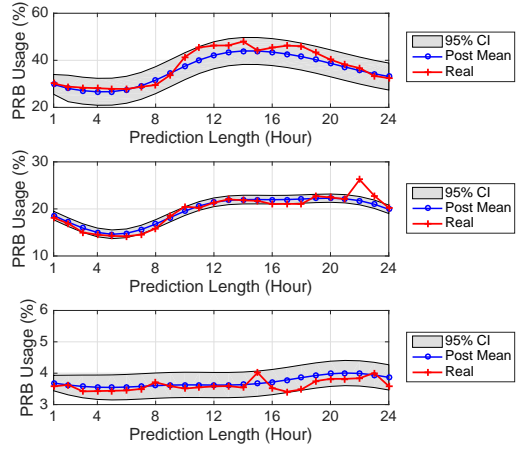}
	}
	\caption{One-hour look-ahead prediction of three RRH clusters. }
	\label{fig:glance}
\end{figure*}
In Fig. \ref{fig:glance}, we present an overview on the prediction results of three RRH clusters, which illustrates the prediction quality in accordance with the wireless traffic patterns presented in Section \ref{sec:patterns}. 
Specifically, for all models in Fig. \ref{fig:prediction}, each time we use 300 data points as the training set, and predict the next one data point (one-hour-ahead). We iteratively update the training set by removing the oldest data point and adding one new data point, and keep going. Since the remained test set contains 420 data points, we can repeat the predictions 420 times for each virtual base station. 
Fig. \ref{fig:prediction} illustrates the prediction results obtained by iteratively updating and predicting 400 times, and each panel corresponds to different types of PRB curves in accordance with the ones were presented in Section \ref{sec:patterns}. 
Specifically, the top and medium panels show that the GP-model, SARIMA and LSTM can capture both daily and weekly patterns, but SS can only capture the daily pattern. The SARIMA model is more easily to be influenced by the bursts in traffic, thus generates a sharp prediction curve, which sometimes can largely degrade the performance, as shown in the bottom panel. The LSTM sometimes overestimates or underestimates the general trend. The GP model usually generates a more robust prediction curve to fit the overall trend. In Fig. \ref{fig:var}, we present the posterior variance obtained by the GP model of a twenty-four-hour-ahead prediction, where the gray area represents the 95\% confidence interval (CI). The result shows that most of the wireless traffic variances fall into the predicted 95\% confidence interval, which indicates that the posterior variance is promising to be used in wireless systems for robust control.

\begin{figure}
	\centering
	\subfigure[]
	{ 	
		\label{fig:mape}
		\includegraphics[trim=1 1 20 1, clip, width=4cm]{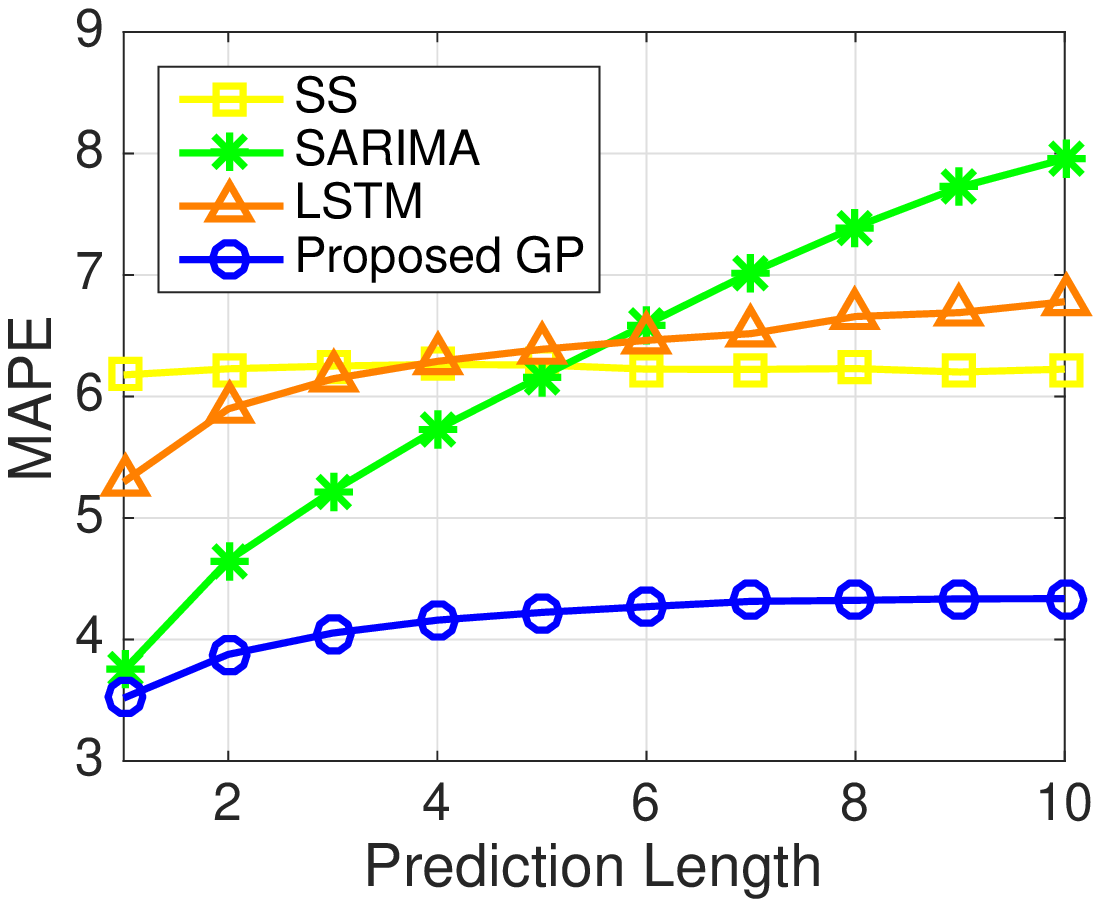}
	}
	\subfigure[]
	{ 	
		\label{fig:rmse}
		\includegraphics[trim=1 1 20 1, clip, width=4cm]{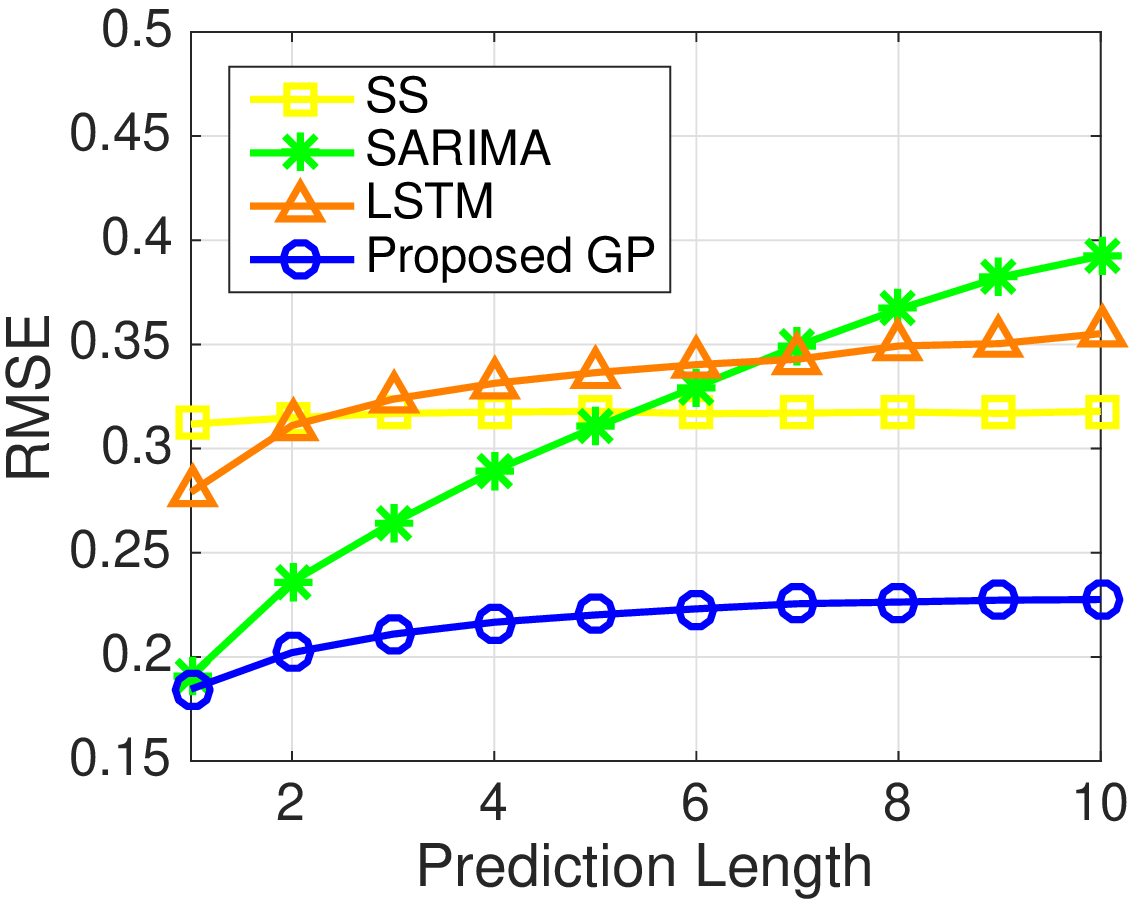}
	}
	\caption{Averaged MAPE and RMSE of downlink PRB usage predictions.}
	\label{fig:performance}
\end{figure}

In Fig. \ref{fig:performance}, we show the averaged RMSE and MAPE of all three models with the prediction length varying from one hour to ten hours. We repeat the prediction 400 times for each cluster and further average the performance over 70 clusters to eliminate the influence due to both temporal and spatial impacts. Therefore, we believe that the obtained performance can present a fair conclusion for general prediction tasks.
Generally, both the RMSE and the MAPE curves show that our proposed GP model can surpass all the competitors. Specifically, the full GP model leads to a prediction error around 3.5\% when predicting the one-hour-ahead traffic, and the error will gradually arise to 4.3\% when the prediction length is extended to 10 hours, then stay stable. 
The SARIMA leads to a prediction error around 4\% at the beginning and increases to 8\% at the end. 
The LSTM leads to a prediction error around 5.2\% at the beginning and 6.8\% at the end. 
Sinusoid superposition has a relatively stable but higher MAPE around 6.2\%.
Note that although the prediction performance of SARIMA is close to that of our proposed model when performing one-hour-ahead prediction, its prediction performance degrades quickly with increased prediction scope. However, it is important to ensure the prediction accuracy on both the short-term and long-term traffic trends, in order to support the real-world traffic-aware managements that aim at maximizing the cumulative utility over a long time range. For example, the traffic-aware RRH on/off operations should be determined according to both the short-term and long-term traffic variations in the future, thereby preventing frequent on/off operations to reduce operational expenditure.

\subsubsection{Performance of Scalable GP}
In the following experiments, we present the comparison between our proposed scalable GP framework and other scalable GP models to demonstrate the superiority of our method.
For each prediction, we use 600 data points as the full training set, and predict the next 10 upcoming data points. We repeat the prediction 100 times for each RRH cluster and further average the performance over 100 RRH clusters to eliminate the influence from random data assignments and abnormal traffic variations. 

\begin{figure}
	\centering
	\includegraphics[trim=20 0 40 0, clip, width=7cm]{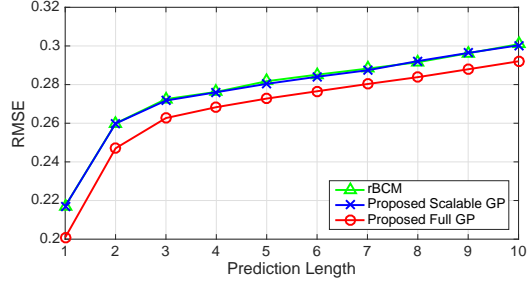}
	\caption{The prediction performance of training phase.}
	\label{fig:training}
\end{figure}
In Fig. \ref{fig:training}, we show the performance of the training phase. The experiments are conducted on three BBUs. In order to compare the performance on the prediction phase solely, for all schemes, we use the same process on the prediction phase as in the full GP, i.e., we use the full dataset for the posterior inference. The curves in Fig. \ref{fig:training} show that both our ADMM based GP and gradient-consensus based rBCM schemes can approximate the training results of the full GP well, with about 9\% performance loss. The result indicates that our proposed ADMM based training scheme can achieve the same performance as the gradient-consensus based rBCM, but with much less communication overheads. Specifically, as discussed in Section \ref{sec:ProposedGPADMM}, the communication overheads required by the gradient-consensus based rBCM and our scheme are $ n_{\text{grads}} * \left( \dim(\boldsymbol{\theta}) + 1 \right) $ and $ n_{\text{cons}} * \left( 2* \dim(\boldsymbol{\theta}) + 1 \right) $, respectively. For each prediction, training the GP model usually requires a few hundreds of gradient steps, i.e., $ n_{\text{grads}} \geq 100 $. Meanwhile, only a few times of ADMM consensuses, i.e., $ n_{\text{cons}} \leq 10 $, is already sufficient for our scheme to converge to an acceptable accuracy level. Therefore, our scheme can save an order of magnitude of communication overheads than the gradient-consensus based rBCM, in general.

\begin{figure}
	\centering
	\includegraphics[trim=20 0 40 0, clip, width=7cm]{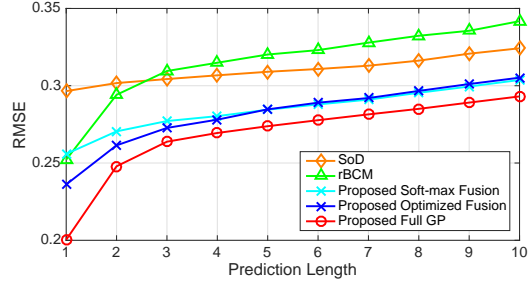}
	\caption{The performance of prediction phase. 1) For our scheme, we set the validation set to be one-point-length, i.e., the closest training point to the test points. 2) For rBCM, we use the differential entropy between the prior and posterior data distribution to be the fusion weights, as in~\cite{deisenroth2015distributed}. 2) For SoD, we randomly choose one-third of the full dataset as the subset to train the sparse GP model, which does not require fusion weights. }
	\label{fig:pred_points}
\end{figure}
In Fig. \ref{fig:pred_points}, we show the performance of the prediction phase. The experiments are conducted on three parallel machines. In this experiment, we bring in the algorithm on the prediction phase for all three schemes. Therefore, on the prediction phase, the scalable GP models no longer have the full dataset; instead, they need to merge the local predictions as the overall result. 
The figure shows that our optimization based algorithm can best approximate the predictions from the full GP, with only 8\% performance loss at the beginning and narrow down to 5\% at the end, almost the same as using the full dataset for prediction.
The soft-max based algorithm is slightly worse than our optimization based algorithm on short-term predictions, but reaches the same level at the latter points. However, both the soft-max based and optimization based fusion schemes can surpass the rBCM and SoD.

\begin{figure}
	\centering
	\subfigure[Performance without concatenation]
	{ 	
		\label{fig:fig_pred_beta}
		\includegraphics[trim=1 0 1 0, clip, width=4cm]{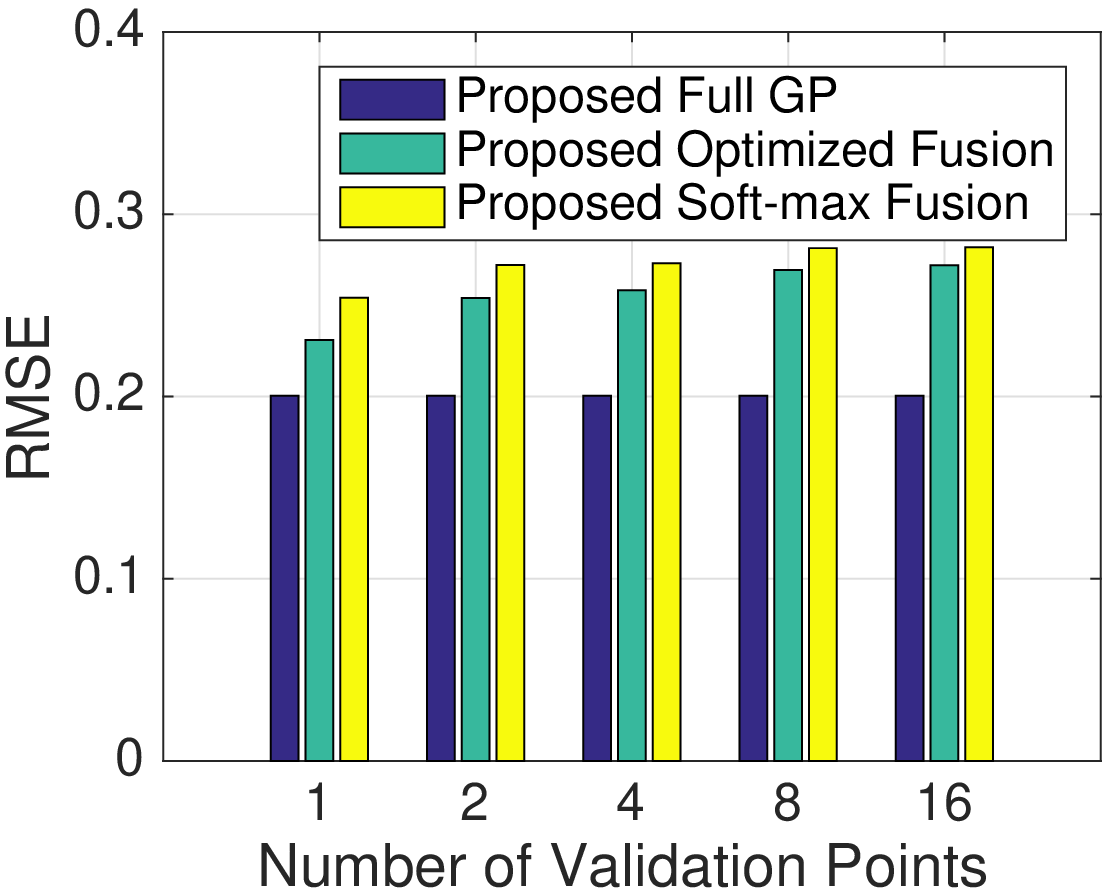}
	}
	\subfigure[Performance with concatenation]
	{ 	
		\label{fig:fig_pred_concat}
		\includegraphics[trim=1 0 1 0, clip, width=4cm]{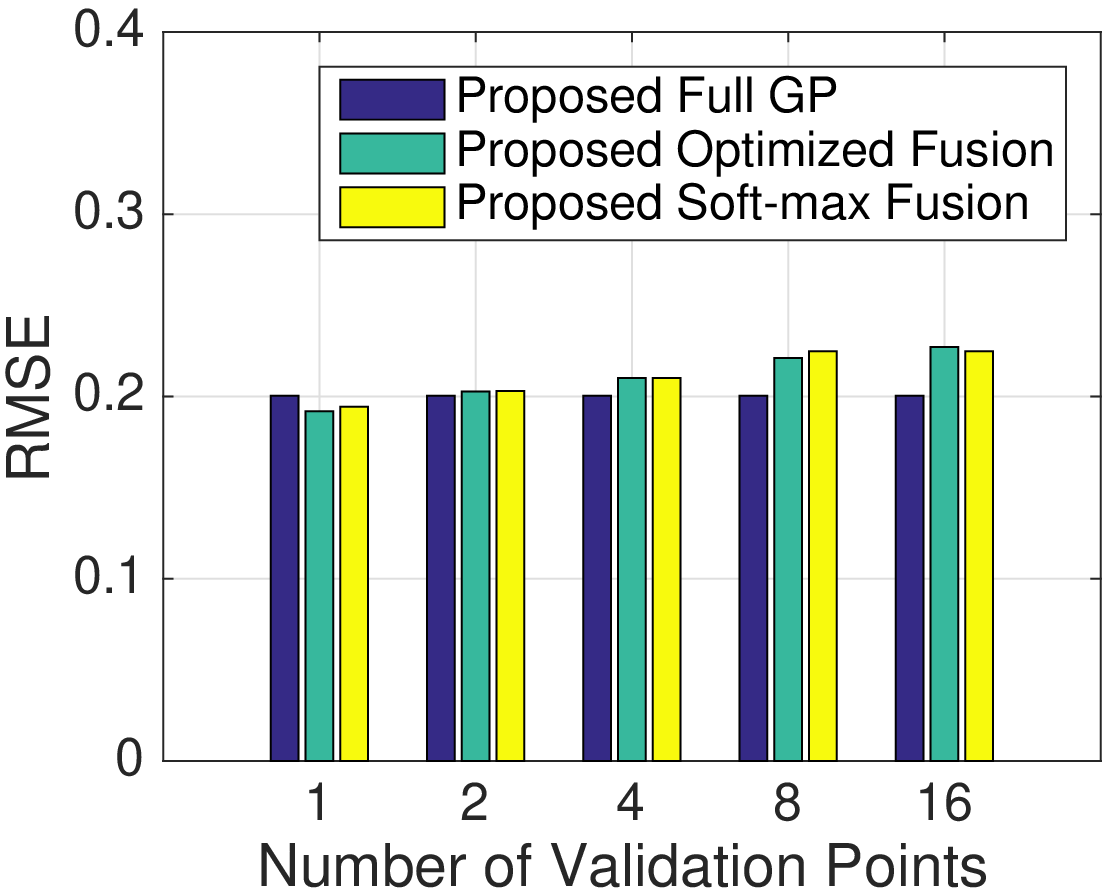}
	}
	\caption{The one-hour-ahead prediction performance.}
	\label{fig:one-hour-ahead performance}
\end{figure}

In Fig. \ref{fig:fig_pred_beta}, we show the one-hour-ahead prediction performance when varying the number of validation points. The experiments are conducted on three parallel machines. The result shows that when the validation set only contains one point, i.e., the closest point to the test points, the prediction model can reach the best performance. The RMSE grows slowly as the validation set size increases. This trend may be due to the fact that our traffic dataset is hourly recorded, such that optimizing the weights on a larger validation set may cause the GP model to ``over-fit" the previous traffic that is far away from the present traffic, while ignoring the most recent changes. 
Currently, we only use the short training set, as shown in Fig. \ref{fig:division}, for posterior inference after determining the fusion weights. However, due to the fact that the validation points contain valuable information about the recent changes close to the test set in regression tasks, the performance could be further improved by using the \textit{full} training set to calculate the posterior mean and variance, which we refer to as the concatenating scheme. 
Specifically, in Fig. \ref{fig:fig_pred_concat}, we show that the concatenating scheme can bring an obvious improvement with all different numbers of validation points.
Moreover, notice that when using one point for validation, the scalable GP with multiple BBUs can even surpass the full GP with one BBU. This is because for certain RRHs, the wireless traffic has many abnormal variations. The full GP uses all the historical traffic data, including those abnormal points, to inference the future traffic. Therefore, the abnormal points with close distance to the test set would have a large influence on the prediction results, which may harm the final predictions. 
However, the scalable GP can filter the bad local inferences based on cross-validation, i.e., by assigning lower weights to poor predictions on the most recent traffic, which results in an improved and robust prediction performance compared to the full GP.
Additionally, it should be pointed out that the optimal validation set size may differ for different datasets. Setting the validation set to be at one-point-length is the optimal choice for our hourly recorded wireless traffic dataset, but may not be so for general cases.

\begin{figure}
	\centering
	\includegraphics[trim=20 0 40 0, clip, width=7cm]{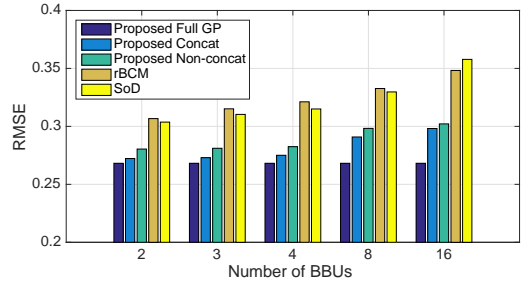}
	\caption{General prediction quality on different number of BBUs. The number of BBUs varies from 2, 3, 4, 8 to 16 with 300, 200, 150, 75, 37 training points, respectively.}
	\label{fig:fig_num_experts_ave}
\end{figure}
In Fig. \ref{fig:fig_num_experts_ave}, we present the general scalability performance by averaging the prediction performance from one-hour-ahead all the way up to ten-hour-ahead for all schemes. The result shows that for all scalable GP models, the prediction performance becomes worse with an increasing number of parallel BBUs, or in other words, with a decreasing number of local training points.
Moreover, the results also show that either with or without the concatenating scheme, our proposed scalable GP model can outperform the rBCM and SoD with considerable performance gains.

\section{Conclusions}
\label{sec:conclusion}
In this paper, we proposed an ADMM and cross-validation empowered scalable GP framework, which could be executed on our proposed C-RAN based wireless traffic prediction architecture. Hence, the framework could exploit the parallel BBUs in C-RANs to meet the large-scale prediction requirements for massive RRHs in a cost-effective way. 
First, we proposed the standard GP model with the kernel function tailored for wireless traffic predictions. 
Second, we extended the standard GP model to a scalable framework. Specifically, in the scalable GP training phase, we trained the local GP models in parallel BBUs jointly with an ADMM algorithm to achieve good tradeoff between the communication overhead and approximation accuracy. 
In the scalable GP prediction phase, we optimized the fusion weights based on cross-validation to guarantee a reliable and robust prediction performance. 
In addition, we proposed the structured GP model to leverage the Toeplitz structure in kernel matrix to further reduce the computational complexity.
The scalable GP framework could easily control the prediction accuracy and time consumption by simply activating or deactivating the BBUs, such that C-RANs could perform a cost-efficient prediction according to real-time system demands.
Finally, the experimental results showed that 1) the proposed GP-based prediction model could achieve better prediction accuracy than the existing prediction models, e.g., the seasonal ARIMA model, the sinusoid superposition model, and the recurrent neural network model; 2) the proposed scalable GP model could achieve the expected goal of complexity reduction while better mitigating the performance loss than the existing distributed GP and sparse GP methods.

\bibliographystyle{IEEEtran}
\bibliography{./bib/IEEEabrv2015.bib,./bib/reference}
\end{document}